\RequirePackage{fix-cm}
\documentclass[a4paper,11pt]{article}

\usepackage{authblk}
\usepackage[english]{babel}
\usepackage[utf8]{inputenc}
\usepackage{microtype}

\newcommand{\R}{\mathbb{R}}

\usepackage{graphicx}%
\usepackage{multirow}%
\usepackage{amsmath,amssymb,amsfonts}%
\usepackage{amsthm}%
\usepackage{mathrsfs}%
\usepackage[title]{appendix}%
\usepackage{xcolor}%
\usepackage{textcomp}%
\usepackage{manyfoot}%
\usepackage{booktabs}%
\usepackage{algorithm}%
\usepackage{algorithmicx}%
\usepackage{algpseudocode}%
\usepackage{listings}%
\usepackage{graphicx}
\usepackage{tabularx}
\usepackage{multirow}
\usepackage{float}
\usepackage{url}
\usepackage{hyperref}
\usepackage{xspace}
\usepackage{todonotes}
\usepackage{paralist}

\newcommand{\GSGP}{$\textsf{GSGP}$\xspace}
\newcommand{\hybrid}{$\textsf{hybrid}$\xspace}
\newcommand{\hybridr}{$\textsf{hybrid}_r$\xspace}
\newcommand{\GPLS}{$\textsf{GPLS}$\xspace}
\newcommand{\GPLSr}{$\textsf{GPLS}_r$\xspace}
\newcommand{\GPLSg}{$\textsf{GPLS}^g$\xspace}
\newcommand{\GPLSrg}{$\textsf{GPLS}_r^g$\xspace}
\newcommand{\regfull}{$\textsf{reg-full}$\xspace}
\newcommand{\regfullr}{$\textsf{reg-full}_r$\xspace}
\newcommand{\reg}{$\textsf{reg}$\xspace}
\newcommand{\regr}{$\textsf{reg}_r$\xspace}
\newcommand{\regg}{$\textsf{reg}^g$\xspace}
\newcommand{\regrg}{$\textsf{reg}_r^g$\xspace}
\newcommand{\gen}{$\textsf{gen}$\xspace}

\tikzset{
    mybrace/.style={decorate,decoration={brace,aspect=#1}}
}

\providecommand{\keywords}[1]{\textbf{\textit{Keywords }} #1}

\begin{document}

\title{Local Search, Semantics, and Genetic Programming: a Global Analysis}

\author[1]{Fabio Anselmi}
\author[2]{Mauro Castelli}
\author[1]{Alberto d'Onofrio}
\author[1]{Luca Manzoni}
\author[3]{Luca Mariot}
\author[1]{Martina Saletta}

\affil[1]{{\small Dipartimento di Matematica e Geoscienze, Università degli Studi di Trieste, Via Alfonso Valerio 12/1, 34127 Trieste, Italy} 

    {\small \texttt{\{fabio.anselmi, alberto.donofrio, lmanzoni\}@units.it, martina.saletta@dia.units.it}}}

\affil[2]{{\small NOVA Information Management School (NOVA IMS), Universidade NOCA de Lisboa, Campus de Campolide, 1070-312 Lisbon, Portugal}

    {\small \texttt{mcastelli@novaims.unl.pt}}}

\affil[3]{{\small Semantics, Cybersecurity and Services Group, University of Twente, Drienerlolaan 5, 7511GG Enschede, The Netherlands} 
	
	{\small \texttt{l.mariot@utwente.nl}}}

\maketitle

\begin{abstract}
Geometric Semantic Geometric Programming (\GSGP) is one of the most prominent Genetic Programming (GP) variants, thanks to its solid theoretical background, the excellent performance achieved, and the execution time significantly smaller than standard syntax-based GP. In recent years, a new mutation operator, Geometric Semantic Mutation with Local Search (GSM-LS), has been proposed to include a local search step in the mutation process based on the idea that performing a linear regression during the mutation can allow for a faster convergence to good-quality solutions. While GSM-LS helps the convergence of the evolutionary search, it is prone to overfitting. Thus, it was suggested to use GSM-LS only for a limited number of generations and, subsequently, to switch back to standard geometric semantic mutation. A more recently defined variant of \GSGP (called \GSGP-reg) also includes a local search step but shares similar strengths and weaknesses with GSM-LS. Here we explore multiple possibilities to limit the overfitting of GSM-LS and \GSGP-reg, ranging from adaptive methods to estimate the risk of overfitting at each mutation to a simple regularized regression. The results show that the method used to limit overfitting is not that important: providing that a technique to control overfitting is used, it is possible to consistently outperform standard \GSGP on both training and unseen data. The obtained results allow practitioners to better understand the role of local search in \GSGP and demonstrate that simple regularization strategies are effective in controlling overfitting.
\end{abstract}


\section{Introduction}
\label{intro}
Genetic Programming (GP)~\cite{koza1994genetic} was introduced 30 years ago as a method to evolve computer programs by mimicking the principles of Darwinian evolution. In its original version, GP uses genetic operators (crossover and mutation) to produce new individuals starting from a given population. These operators allow the creation of new solutions by operating on the syntax of the existing solutions. In other words, both crossover and mutation generate new children by acting on the structure of the parent solutions. In the most common implementation of tree-based GP, crossover produces two children by swapping parents' subtrees, while mutation replaces a parent's subtree with a randomly generated one.

While research demonstrated the suitability of standard syntax-based GP in addressing complex problems over different domains and reaching human-competitive results~\cite{koza2010human}, researchers started to analyze the properties of the evolutionary process. From this analysis, an important drawback related concerning syntax-based genetic operators emerged, despite their implementation simplicity. In particular, knowing the outputs of the parents is not useful for predicting the outputs of the children. This is due to the fact that the standard, syntax-based genetic operators work by only considering the structure of the parents and ignore the results they produced over the available observations~\cite{vanneschi2014survey}.
Even more important, the output of a GP solution is what matters to practitioners. Thus, it is clear that genetic operators which blindly modify the parent individuals without knowing the effect of their modification on the children's outputs represent a relevant limitation for standard GP.

To overcome this limitation, GP research started investigating methods to include semantic awareness in the evolutionary process. 
Krawiec and Pawlak~\cite{krawiec2013locally} proposed the definition of semantics nowadays used in the GP literature by taking inspiration from the field of formal languages, where semantics describes the effects of program execution for all possible input data. GP evaluates programs over a finite training set of fitness cases, each being a pair consisting of an input and the corresponding desired program output. Thus, assuming that this set is the only available data to specify the output of the sought solution, a natural definition of semantics---which we consider in this paper---is the vector of outputs a program produces for a set of observations~\cite{vanneschi2014survey}. By adopting this definition, the semantics of an individual is a point in an $n$-dimensional space (i.e., the semantic space), where $n$ is the number of fitness cases.

The first studies trying to include semantic awareness in the evolutionary process were based on a simple idea: the use of traditional syntax-based genetic operators followed by the evaluation of some semantic criteria on the resulting children to accept or reject the newly generated solutions~\cite{vanneschi2014survey}. Despite the satisfactory results achieved by these methods, they presented significant shortcomings. First, they were time-consuming due to the effort needed to evaluate the semantic criteria, and second, a high number of individuals were rejected. These limitations made the evolutionary process unbearably slow, thus limiting the use of this semantic criteria only to toy problems.

The situation changed in 2012 when Moraglio and coauthors~\cite{moraglio2012geometric} proposed a method allowing the inclusion of semantic awareness in a direct manner. In particular, they defined genetic operators called geometric semantic operators (GSOs) that can directly act on the semantics of the parent individuals to produce new solutions. Additionally, GSOs induce a unimodal fitness landscape on any supervised problem (i.e., regression and classification)~\cite{moraglio2004topological}.
The resulting GP algorithm was called geometric semantic genetic programming (\GSGP).
\GSGP is nowadays a hot topic in evolutionary computation, with many studies showing its ability to outperform syntax-based GP over different domains~\cite{castelli2017evolutionary,castelli2017predicting,castelli2015prediction}.
Although existing implementations~\cite{castelli2015c++,CASTELLI2019100313} of \GSGP allow for a faster evaluation of a candidate solution than syntax-based GP, the number of iterations needed to converge and the increasing size (i.e., number of nodes) of the \GSGP trees make it difficult to use \GSGP to address problems characterized by a vast amount of data, and when the interpretability of the model is a fundamental requirement.

A promising path to tackle this issue is the combination of \GSGP with a local search strategy~\cite{castelli2015geometric}. The inclusion of a local search method within \GSGP aims at speeding up the search process so that good-quality solutions are obtained in a limited number of iterations. Moreover, running \GSGP for a limited number of iterations allows to obtain smaller solutions.
Local search methods coupled with GP represent an important research area in evolutionary computation, and this paper continues this investigation focusing on \GSGP. In particular, this work is the natural consequence of two studies in which a local search optimizer was coupled with \GSGP: in the first study~\cite{castelli2015geometric}, the authors modified the geometric semantic mutation operator to incorporate a local search operator, resulting in a new mutation called GSM-LS; in the second study~\cite{EPIA} the authors presented promising results achieved by applying a new local search operator (\GSGP-reg) to all the individuals during a separate step after mutation and crossover. 
Despite the satisfactory results achieved in both studies, \GSGP with local search may result in severe overfitting in some of the considered problems.
Thus, in this research, we investigate a strategy to limit the overfitting of GSM-LS and \GSGP-reg, ranging from adaptive methods to estimate the risk of overfitting at each mutation to a simple regularized regression. We will compare two techniques to include local search in \GSGP and present results that may help practitioners to use \GSGP with local search and how to limit overfitting through regularization.

The paper is organized as follows: Section~\ref{sec:statoarte} recalls the definition of GSOs and previous work on local search methods within \GSGP; Section~\ref{sec:ourwork} describes the strategy proposed in this paper to counteract overfitting; Section~\ref{sec:settings} outlines the benchmark problems considered in this study and the experimental settings; Section~\ref{sec:results} discusses the results by comparing the performance of different local search methods. Finally, Section~\ref{sec:conclusions} concludes the paper by summarizing the main contributions of this work.

\section{Background}
\label{sec:statoarte}

In this section, we first recall the definitions of the geometric semantic operators used in this work. Subsequently, we outline existing approaches to integrate local search in standard GP and \GSGP. 

\subsection{Geometric Semantic Operators}

In 2012, Moraglio and coauthors~\cite{moraglio2012geometric} introduced Geometric Semantic Operators (GSOs), genetic operators that directly incorporate semantic awareness in the evolutionary process. Such GSOs produce a transformation on the syntax of the individuals that correspond to geometric crossover and ball mutation~\cite{krawiec2009approximating} in the semantic space. More in detail, the geometric crossover generates only one child that stands, in the semantic space, between the two parent solutions. On the other hand, the geometric semantic mutation produces a new solution whose coordinates in the semantic space correspond to a slight perturbation of some of the coordinates of the parent solution. 

In this study, we will focus on regression problems. Thus, we report the definition of the GSOs for real function domains. For further details of GSOs in different domains, the reader is referred to~\cite{moraglio2012geometric}.

\textbf{Geometric Semantic Crossover (GSC)}.
Given two parents represented as functions $T_1, T_2 : \mathbb{R}^n \to \mathbb{R}$, the geometric semantic crossover returns the real function $T_{XO}=(T_1 \cdot T_R) + ((1-T_R)\cdot T_2)$, where $T_R$ is a random real function whose output values range in the interval $[0,1]$. As a results, for each $x \in \mathbb{R}^n$, if $T_1(x) \le T_2(x)$ then $T_1(x) \le T_{XO} (x) \le T_2(x)$.

\textbf{Geometric Semantic Mutation (GSM)}.
Given a parent function $T : \mathbb{R}^n \rightarrow \mathbb{R}$, the geometric semantic mutation with mutation step $ms$ returns the real function $T_M=T+ms\cdot(T_{R1} - T_{R2})$, where $T_{R1}$ and $T_{R2}$ are random real functions. 

As shown in~\cite{castelli2016semantic,enriquez2017automatic}, limiting the codomain of $T_{R1}$ and $T_{R2}$ to $[0,1]$ helps improving the generalization ability of \GSGP. With this constraint, for all $x \in \mathbb{R}^n$, we have that $|T(x) - T_M(x)| \le ms$, i.e., $T_M$ is a point in the semantic space that lies on a ball of radius $ms$ centered in $T$.

\subsection{Local Search and GP}

Coupling evolutionary algorithms (EAs) with local search strategies has been investigated in different studies~\cite{chen2011multi,neri2012memetic,vcrepinvsek2013exploration}. The fundamental idea shared by these methods (often called memetic algorithms) is to fully explore the local region around each individual. 
By exploiting this principle, memetic algorithms were shown to achieve a better performance over the standard evolutionary approach~\cite{neri2012memetic,chen2011multi}, outperforming evolutionary algorithms also in multimodal optimisation~\cite{nguyen2020memetic}.
Despite these promising results, the literature features a relatively low number of contributions that couple GP with local search~\cite{trujillo2018local}. In particular, the two main approaches to integreate a local search (LS) strategy into GP are: (1) apply LS on the syntax~\cite{azad2014simple,eskridge2004imitating}; or (2) apply it on numerical parameters of the program~\cite{topchy2001faster,zhang2004genetic}.
In both cases, existing contributions showed that the strategy used to apply local search in GP has a significant impact on the performance of the resulting model. Focusing on the use of GP to address symbolic regression problems, Z-Flores and coauthors~\cite{emigdio2014evaluating} suggested that the best strategy consists in the application of LS to all the individuals in the population or, eventually, to a subset of the best individuals. Moreover, the same study empirically showed that including an LS strategy improves convergence and performance while reducing the growth of the solutions' size.

Focusing on \GSGP, the integration of a local search strategy in the search process is a recent topic. To the best of our knowledge, the first contribution in this research strand was proposed by Castelli and coauthors~\cite{castelli2015geometric} in 2015, where a greedy LS method was integrated into the geometric semantic mutation operator. In particular, given a parent individual T, the mutation operator (GSM-LS) is defined as follows: 

\[
  T_M = \alpha_0 + \alpha_1 \cdot T + \alpha_2 \cdot (R_1 - R_2)
\]
where $R_1$ and $R_2$ are random trees with output in $[0,1]$, $\alpha_i \in \R$.

As reported by Castelli and coauthors~\cite{castelli2015geometric}, the GSM-LS operator aims at determining the best linear combination of the parent tree and the random trees (R1 and R2), and it is local in the sense of the linear problem it defines.
The GSM-LS operator was considered in different studies ~\cite{castelli2015energy,hajek2019forecasting,trujillo2018local} and demonstrated its suitability in speeding up the convergence of the search process, thus reducing the size of the resulting solution. However, severe overfitting appeared in some of the investigated benchmarks, suggesting that further research is needed to fully understand how LS can improve \GSGP's performance without hampering its generalization ability.

The initial proposal presented in~\cite{castelli2015geometric} was subsequently extended in~\cite{EPIA}, where the authors introduced a new local search operator (\GSGP-reg, or \reg for short) by defining a generic set of functions that locally modify a candidate GP individual. The main idea is that the possibility of modifying the set of functions allows the creation of a problem-specific local search operator that should improve \GSGP's performance. \GSGP-reg achieved a performance comparable to \GSGP-LS. However, as the authors pointed out, the problems in which \GSGP-reg and \GSGP-LS tend to overfit are different, thus suggesting the need for further investigation in this area.

Recently, Pietropolli and coauthors~\cite{pietropolli2022combining} proposed the use of an LS optimizer (i.e., Adam~\cite{kingma2017adam}) that acts on both GSOs introduced in~\cite{moraglio2012geometric}. Their idea is the following: after performing the evolutionary steps involving GSM and GSC, a new population $T=(T_1, T_2, \cdots, T_M)$ of $M$ individuals is obtained, with $T$ composed of differentiable functions (as they are obtained through additions, multiplications, and compositions of derivable functions).
By considering of GSC defined as $T_{XO} = (T_1 \cdot \alpha) + ((1 - \alpha) \cdot T_2)$ (with $0 \le \alpha \le 1$) and GSM as $T_M = T + ms \cdot (R_1 - R_2)$ (with $0 \le ms \le 1$), and considering that the values of $\alpha$ and $ms$ are randomly initialized, it is possible to derive $T$ with respect to $\alpha$, $\beta=(1-\alpha)$, and $m$. 
Therefore, the gradient-based optimizer can be applied considering $\theta=(\alpha, \beta, ms)$ as the parameters vector. 
While the resulting algorithm can be considered as a local search, it deviates from GSM-LS and \GSGP-reg by performing non-local modifications to the tree representation. In fact, not only the parameters to be optimized modify the (parameters of) crossovers and mutations, but, due to the representation of \GSGP trees as a directed acyclic graph, the optimization is performed on the entire population at the same time. Due to this significant difference, we will not consider this approach in this work.

\section{Adaptive Local Search}
\label{sec:ourwork}

Since one of the main problems of local search in \GSGP is the risk of overfitting, a reasonable approach would be to test whether a local search step produces a solution that generalizes well before accepting it. If the assessment shows poor generalization ability, the local search step is rejected, and the original individual is returned.
The information concerning the fraction of accepted solutions obtained with the LS application can be used to change the probability of applying it. That is, if most local search steps result in overfit individuals, it would be better to reduce the probability of applying a local search step.
Those two principles are encapsulated in the ``\gen'' adaptive local search algorithm presented below.

Let us consider a \GSGP tree $T$ evaluated on a training set $X$ consisting of $m \in \mathbb{N}$ individuals. Let $\ell$ be a local search operator returning a tree $T' = \ell(T; X)$; notice that the local search step might depend on the training set $X$ that can be employed in building $T'$, as in GSM-LS and \GSGP-reg. From now on, the fitness of $T$ on the set $X$ will be denoted as $f(T; X)$.

This general idea of the adaptive local search mechanism \gen (the name stems from generalization-based) is encapsulated in two algorithms. The first one accepts the local search only if it is expected to give better fitness on unseen data as follows:
\begin{enumerate}
    \item Split the training set $X$ into two non-overlapping subsets $X_1$ and $X_2$;
    \item Apply the local search operator $\ell$ on $X_1$ to obtain $T'=\ell(T; X_1)$;
    \item If $f(T';X_2)$ is improved with respect to $f(T;X_2)$ then $T'$ is returned;
    \item Otherwise, $T$ is returned.
\end{enumerate}
We expect that if $\ell$ is applied and the tree $T'$ is accepted, then $T'$ will perform well on unseen data, thus avoiding overfitting. If, instead, $T$ is returned, we expect $T'$ to overfit.
In our current implementation, we set $|X_1| = \lceil 0.9|X| \rceil$ and $|X_2| = \lfloor 0.1|X| \rfloor$.

Notice that some additional information can be obtained from multiple applications of the previous algorithm.
If the individuals produced by the local search are rejected with a high frequency, the current population may be in a part of the search space where the local search operation $\ell$ is not beneficial. Hence, the second component of the \gen algorithm is to apply the local search to each tree with a certain probability.

In particular, let $t$ be the current generation and let $N_\text{acc}(t-1)$ be the number of accepted local search operations (i.e., local search operations resulting in the acceptance of $T'$) up to generation $t-1$ over a total of $N(t-1)$ tries. Then, at the current generation, the probability of performing a local search for a tree is $\max \left \lbrace 0.01, \frac{N_\text{acc}(t-1)}{N(t-1)}\right\rbrace$. Notice that there is a minimum probability of $0.01$ for the local search to be applied, thus avoiding the removal of the local search altogether.

Now, let $n_\text{acc}(t)$ be the number of accepted local search operations for generation $t$ and $n(t)$ the number of local search operations executed at generation $t$. Then, the cumulative number of accepted and executed local search operations up to generation $t$ are respectively $N_\text{acc}(t) = N_\text{acc}(t-1) + n_\text{acc}(t)$ and $N(t) = N(t-1) + n(t)$. This means that the resulting probability will progressively change less, because the influence that a generation can have on its variation decreases over time.

\section{Experiments}
\label{sec:settings}

In this section, we define the objective of our experiments, the underlying settings and datasets, and the methods considered in our investigation.

First of all, there are three main research questions that we want to address in the experimental phase:

\begin{itemize}
    \item[\textbf{RQ1}] Are the overfitting control strategies for local search able to prevent overfitting?
    \item[\textbf{RQ2}] Are these strategies still able to provide better performances than plain \GSGP?
    \item[\textbf{RQ3}] Is there one strategy that can provide better performances than the others?
\end{itemize}

To answer these research questions, we will compare the performances of the different algorithms on seven different regression problems.
In particular, for RQ3, we will consider both a very crude solution, i.e., using local search only in the first few generations, the adaptive search algorithm \gen, and the use of linear regression with $L_2$ regularization in the local search step.

\subsection{Tested Algorithms}

Three baseline algorithms are considered and, for each of them, one or more overfitting-preventing techniques are applied, namely:
\begin{itemize}
    \item Use the local search method---either GSM-LM or \reg---only for the first ten generations. This is the original approach proposed for both techniques;
    \item Application of the \gen adaptive local search method;
    \item Use ridge regression~\cite{marquardt1975ridge} instead of the least squares linear regression with no regularization. In ridge regression, an additional term provides an~$L_2$ regularization to avoid overfitting.
\end{itemize}
The addition of ridge regression as an overfitting prevention method is because overfitting---and modeling noise or outliers in the data---is a risk even for a simple linear regression. Thus, the use of a regularization factor limits the risk of overfitting.

We consider \GSGP, \GPLS (i.e., \GSGP with the GSM-LS operator), and \reg as the baseline algorithms. A combination of these algorithms with the different overfitting prevention strategies leads to assess the following variants:
\begin{itemize}
    \item \GSGP. This is the standard \GSGP where no local search step is performed.
    \item \GPLS. In this case, for all the generations, the mutation operator is GSM-LS.
    \item \GPLSr. The regression used inside GSM-LS is ridge regression.
    \item \GPLSg. The mutation operator GSM-LS uses the adaptive search \gen.
    \item \GPLSrg. In this case, the mutation operator GSM-LS uses the adaptive search \gen, and the regression employed is ridge regression.
    \item \hybrid. Like \GPLS but the GSM-LS operator is used only in the first $10$ generations.
    \item \hybridr. Like \GPLSr but the GSM-LS operator is applied only for the first $10$ generations.
    \item \regfull. In this case, the local search defined in \reg is applied in all generations.
    \item \regfullr. Like \regfull but the regression used is ridge regression.
    \item \reg. The originally defined \reg where the search is applied only in the first $10$ generations.
    \item \regr. Like \reg but ridge regression is employed.
    \item \regg. Like \reg but using \gen as an overfitting prevention technique.
    \item \regrg. The combination of \regr (use of ridge regression) and \regg (use of \gen) at the same time.
\end{itemize}
As one can see, a subscript $r$ indicates the use of \emph{ridge regression}, while a superscript $g$ represents the use of the \gen algorithm.

\subsection{Experimental Settings}

For all tested methods, the set of functional symbols is $\{+, -, \times, \div \}$, where $\div$ is protected by returning $1$ when the denominator is sufficiently close to $0$. The set of terminal symbols is the set of input variables of the problems used in the benchmarks, with no fixed constants belonging to the terminal set. The trees in the initial generation and the random tree used by the semantic operators were all generated with a ramped-half-and-half initialization method with a maximum depth of $6$. During the evolution, the selection method used was tournament selection with a tournament size of $4$, while mutation and crossover were performed -- in an exclusive way -- with probabilities of $0.6$ and $0.4$, respectively. In the case of the \gen method applied to GSM-LS, the two probabilities were combined multiplicatively (i.e., once an individual has been selected for mutation, it can either be subject to a local search step or remain unchanged). For all methods elitism is used, with the best performer on the training set at one generation surviving to the next one. For the methods based on \reg, the function used in the regression were the tree $f_1(x) = T(x)$, $f_2(x) = 1$, $f_3(x) = \min\{0, T(x)\}$ (i.e., the negative part), and $f_4(x) = \max\{0, T(x)\}$ (i.e., the positive part), with $T(x)$ representing the tree on which the local search is to be applied evaluated on input $x$. When ridge regression is used, the regularization factor is set to $0.001$; tests with values of $0.01$ and $0.0001$ were also performed with no change in the trends of the results. Thus, to simplify the discussion, only the results for $0.001$ are shown.

The population size is set to $100$ individuals, evolved for $100$ generations. For each dataset, $100$ independent runs are performed with a different random split between training ($70\%$ of the samples) and test sets ($30\%$ of the samples). In each run, the fitness on the training and test set of the best individual, computed as the root mean squared error (RMSE), was recorded for each generation. For all the methods using the \gen algorithm, the probability of performing a local search step is also recorded at each generation. To compare the results at the last generation, the Mann-Whitney U-test (with Bonferroni correction) is used since it makes no assumption on the distribution of the underlying samples.

\subsection{Datasets}

To perform a comparison of the different methods, we used $7$ different symbolic regression problems, with some of them widely used in the community~\cite{mcdermott2012genetic,white2013better}. They are all real-world data collected from multiple domains, as detailed below:
\begin{itemize}
    \item \textit{Airfoil}. The dataset consists of different airfoils at various wind tunnel speeds and angles of attack, with the target being the scaled sound pressure level in decibels.
    \item \textit{Human Oral Bioavailability} (bioav). This dataset has a pharmacokinetic parameter as a target, which measures the quantity of orally-administered drug that enters the blood circulation after being processed by the liver. Each instance consists of $241$ molecular descriptors.
    \item \textit{Concrete Compressive Strength} (concrete). The target is a parameter measuring the resistance to compression of a particular mix of concrete. Each instance is a different concrete mix described by $6$ variables.
    \item \textit{Parkinson}. The dataset consists of voice measurements of $42$ patients in the early stage of Parkinson's disease recruited for a six-month trial of symptom progression monitoring. The input variables consist of patient information and $16$ voice measurements. The target is to predict the total UPDRS (Unified Parkinson's Disease Rating Scale) score.
    \item \textit{Plasma Protein Binding} (ppb). Plasma Protein Binding is a parameter measuring the quantity of drug that reaches circulation and further attaches to plasma proteins in the blood. The dataset is composed of $131$ molecules instances, each described by $626$ features.
    \item \textit{Concrete Slump} (slump). The target variable, i.e., concrete slump, is a parameter that measures the consistency of fresh concrete. Each instance is a concrete mix described by $8$ variables.
    \item \textit{Median Letal Dose} (LD50). Median Lethal Dose measures the quantity of a drug necessary to kill half of the test organisms. This dataset is obtained by considering mice as test organisms and oral supplying as an administration route. Each instance represents a molecule described by $626$ features.
\end{itemize}

The number of instances and the number of variables for each dataset are summarized in Table~\ref{tab:datasets}. For more details on the datasets, we refer the reader to~\cite{castelli2013prediction,yeh1998modeling} for the concrete and slump problems, to~\cite{archetti2007genetic} for all the pharmacokinetic datasets, to~\cite{tsanas2009accurate} for the Parkinson dataset and to~\cite{brooks1989airfoil} for the airfoil problem.

\begin{table}[t]
    \renewcommand{\arraystretch}{1}
    \begin{tabular}{rrrrrrrr}
        \toprule
        & airfoil & bioav & concrete & parkinson & ppb & slump & LD50 \\
        \midrule
        \textit{\# variables} & $5$ & $241$ & $8$ & $18$ & $628$ & $9$ & $6$ \\
        \textit{\# instances} & $1502$ & $359$ & $1029$ & $5875$ & $131$  & $102$ & $307$ \\
        \bottomrule
    \end{tabular}
    \caption{\label{tab:datasets} Number of input variables and instances for each dataset used in the experimental phase.}
\end{table}

\section{Results}
\label{sec:results}

\begin{figure}
    \centering
    \renewcommand{\arraystretch}{0}
    \begin{tabular}{c@{}c}
        {\footnotesize airfoil} & {\footnotesize bioav} \\ 
        \includegraphics[width=0.48\textwidth]{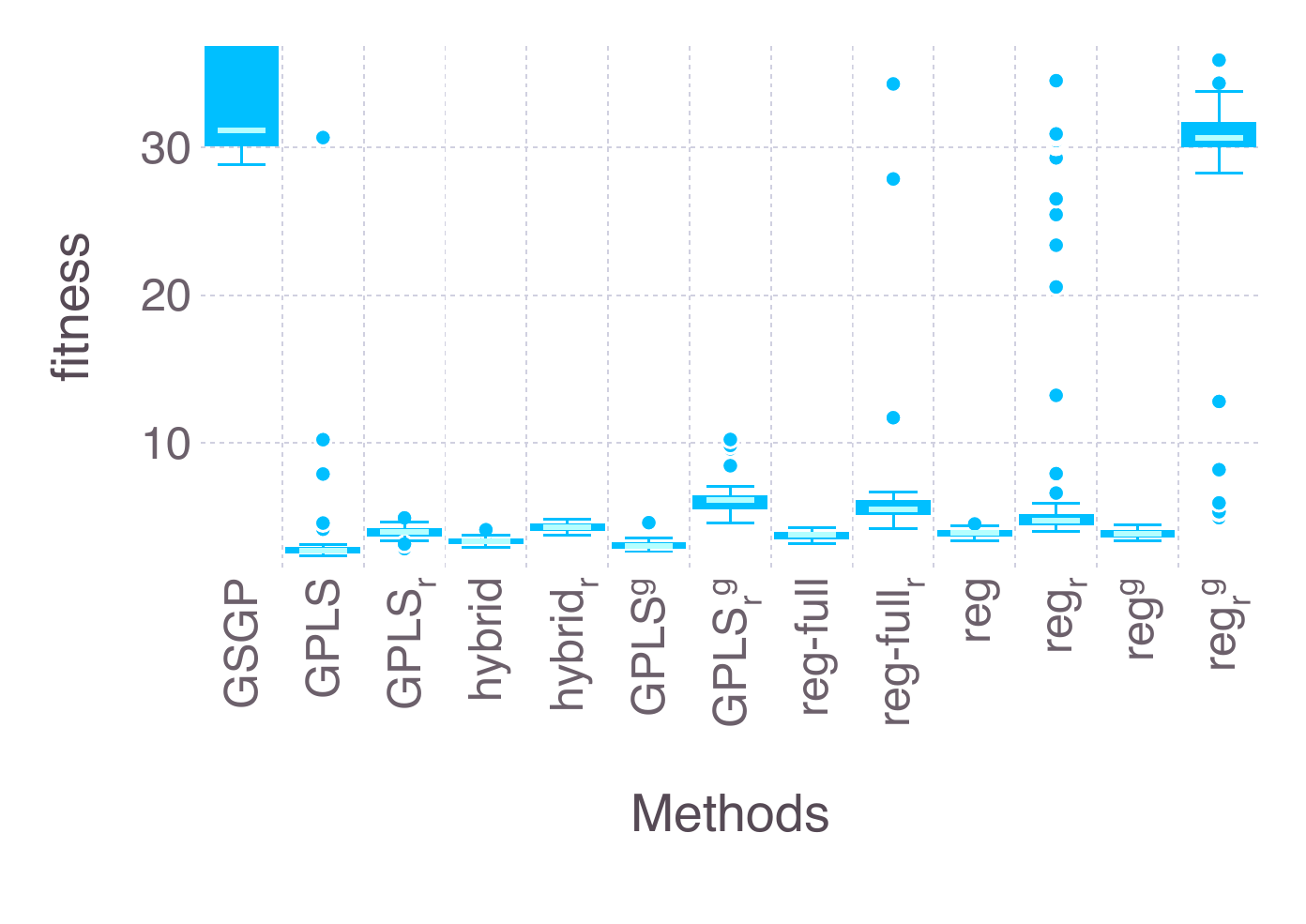} &
        \includegraphics[width=0.48\textwidth]{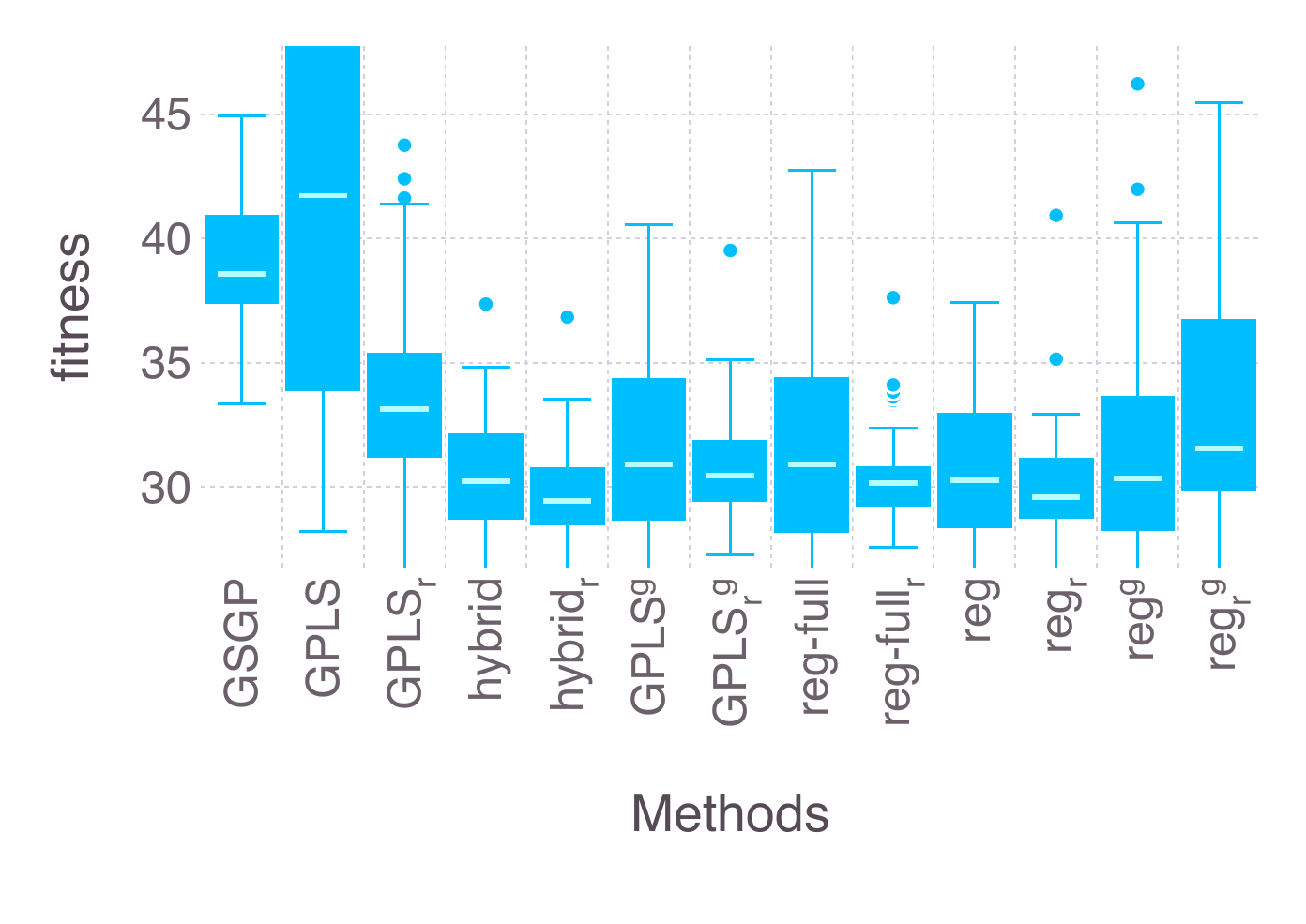} \\
        {\footnotesize concrete} & {\footnotesize parkinson} \\
        \includegraphics[width=0.48\textwidth]{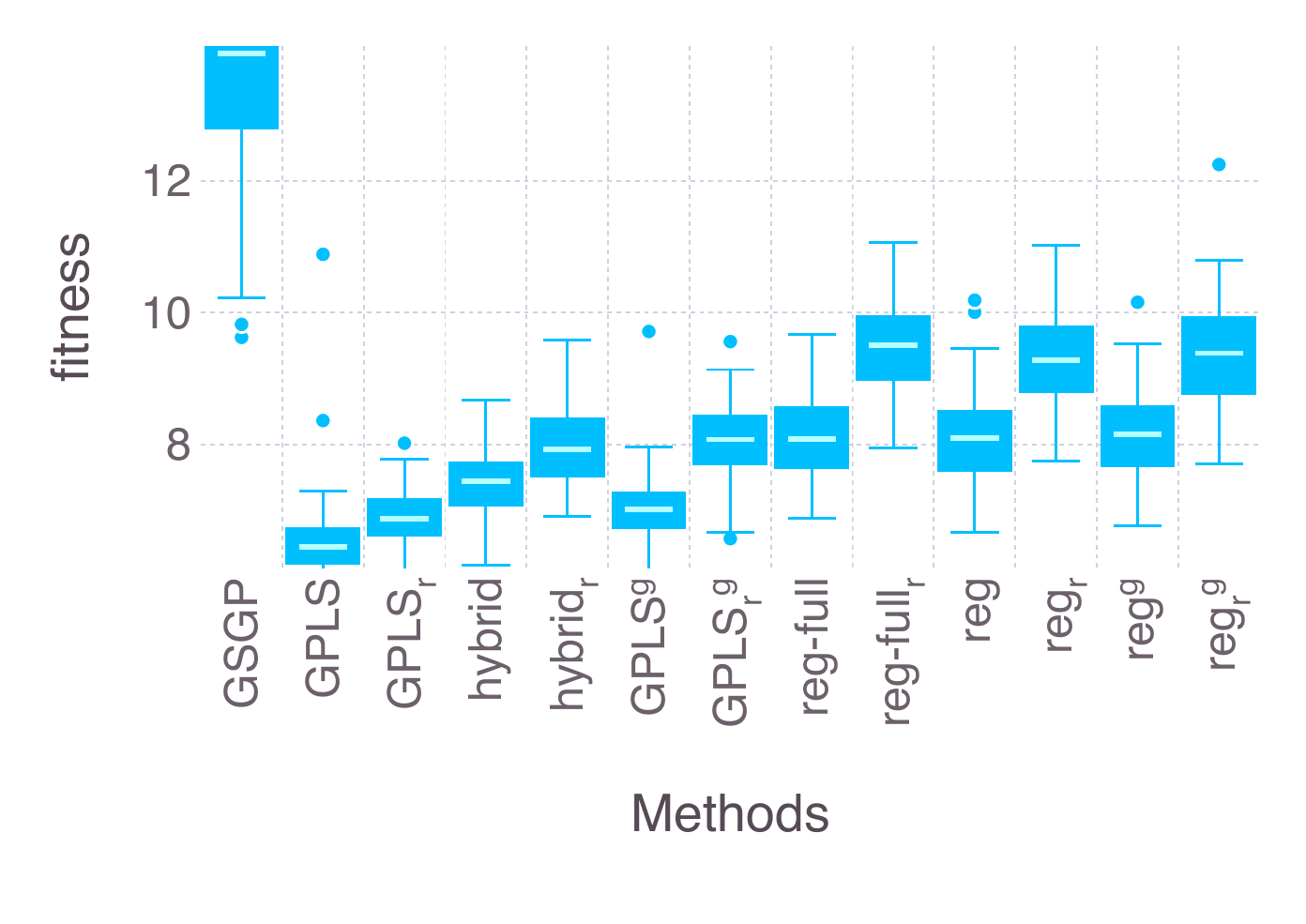} &
        \includegraphics[width=0.48\textwidth]{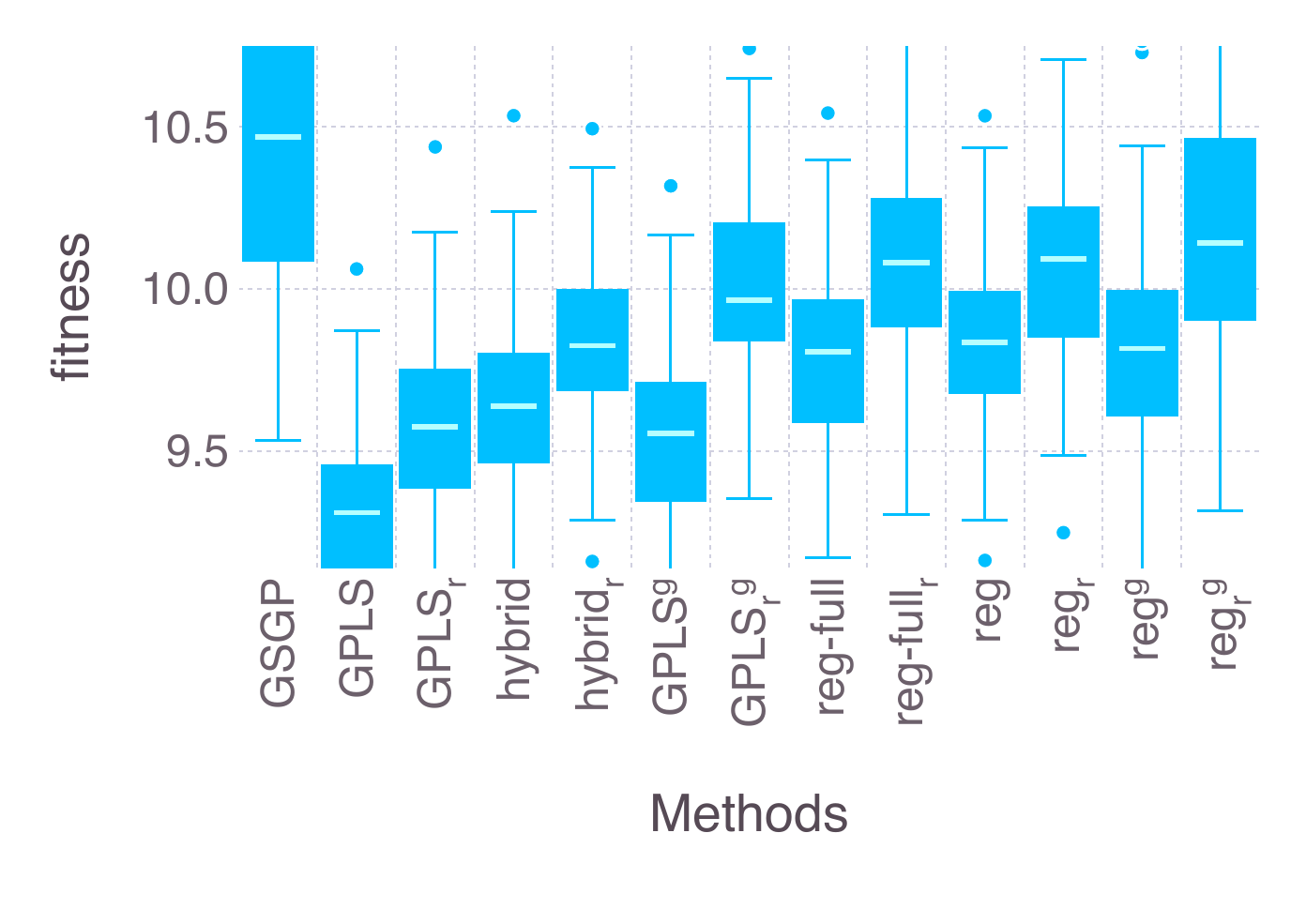} \\
        {\footnotesize ppb} & {\footnotesize slump} \\
        \includegraphics[width=0.48\textwidth]{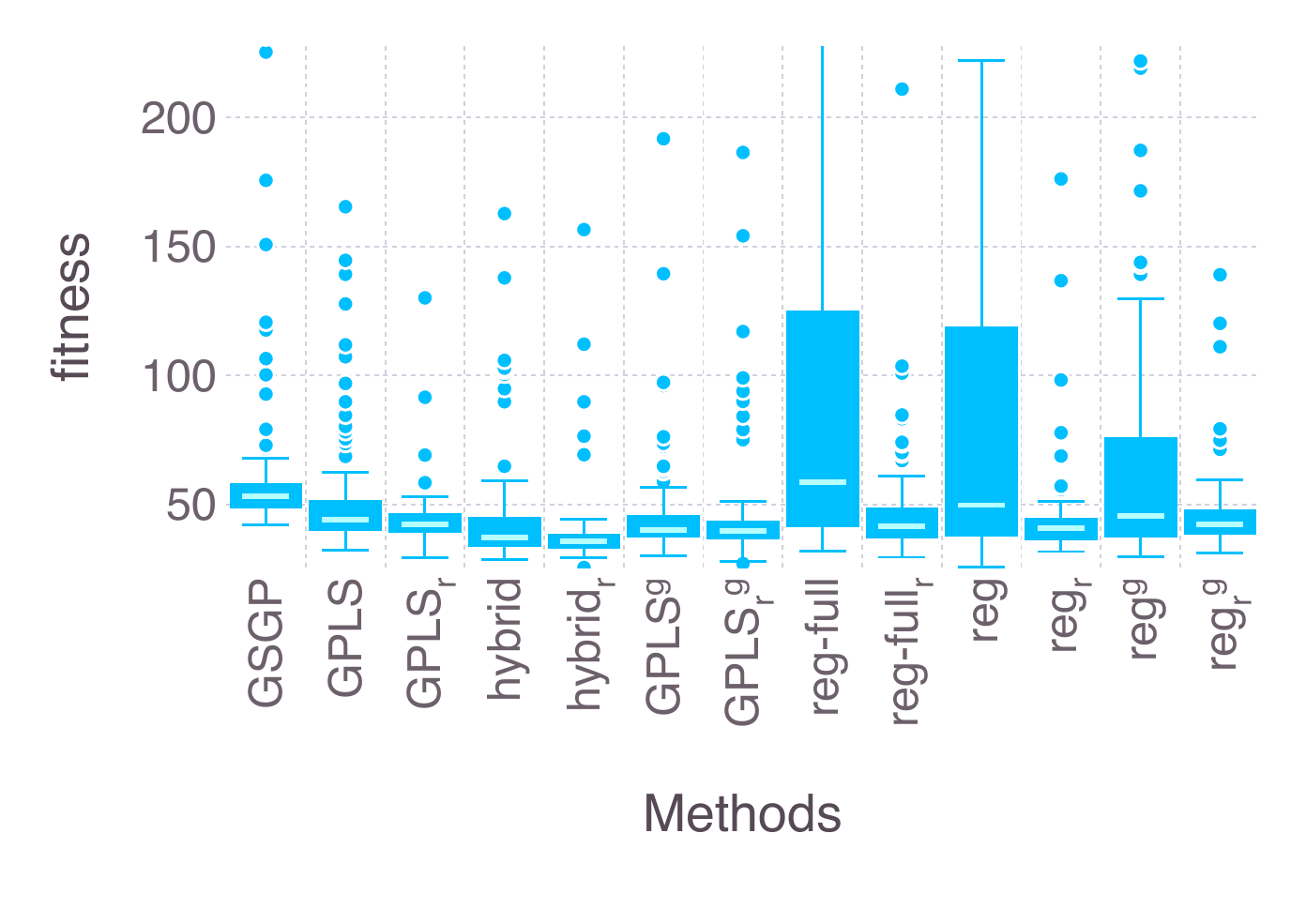} &
        \includegraphics[width=0.48\textwidth]{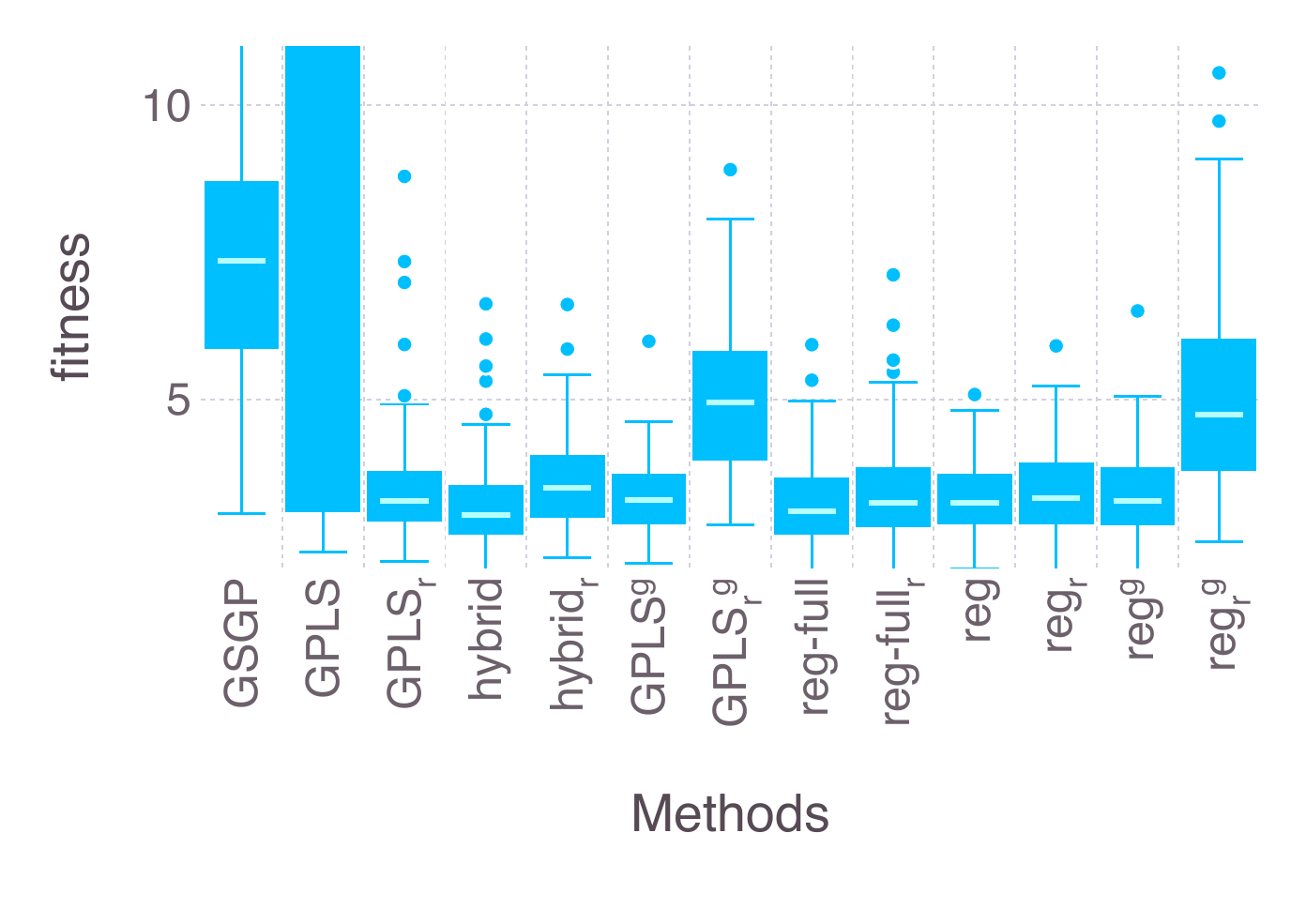} \\
        {\footnotesize LD50} & \\
        \includegraphics[width=0.48\textwidth]{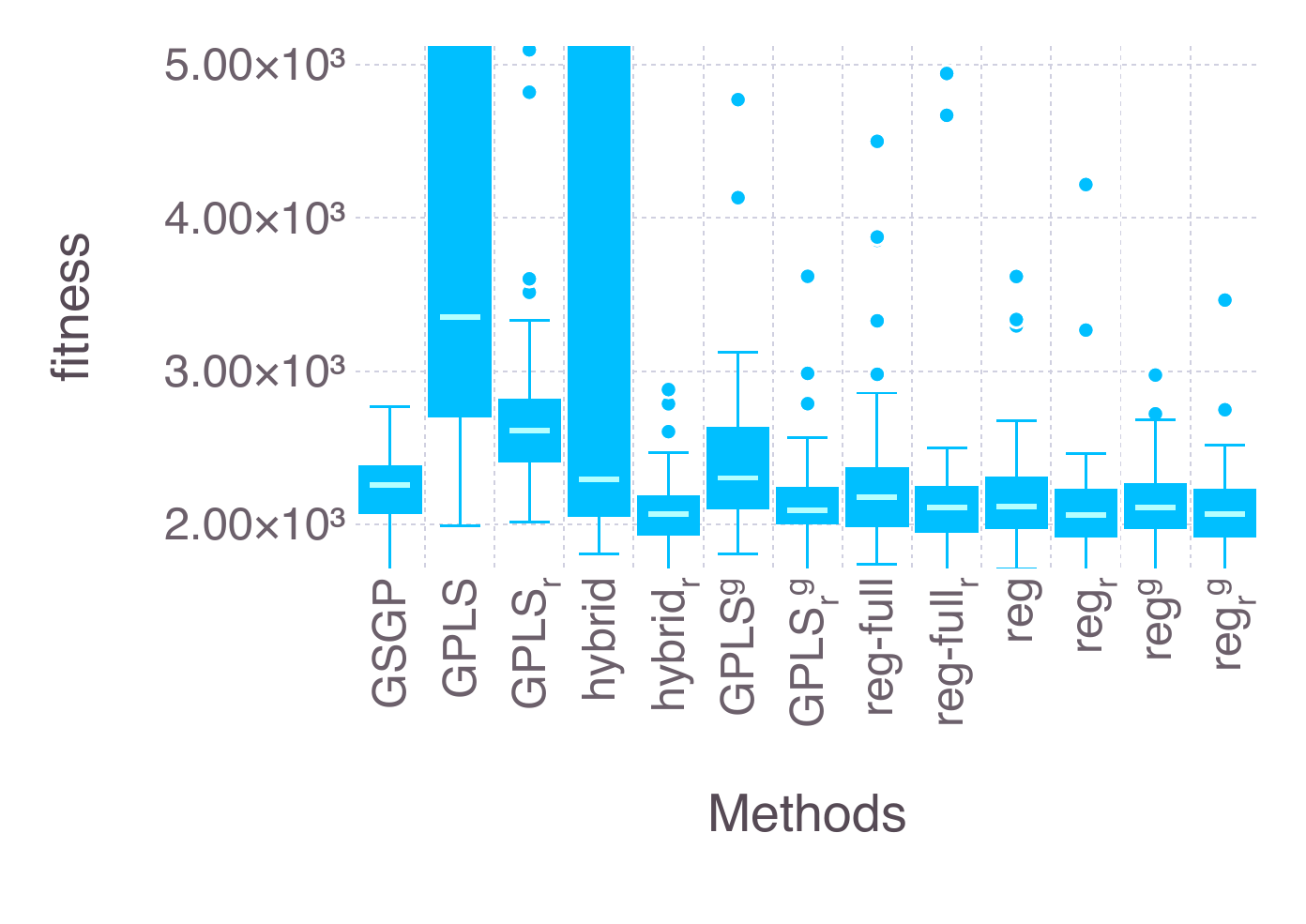} & \\
    \end{tabular}
    \caption{Boxplots of the fitness at the last generation on the \emph{test set} for all the considered problems. The boxes show the second and third quartiles, with the internal line showing the median. The whiskers show the first and fourth quartiles, excluding the outliers that are explicitly shown as points.}
    \label{fig:boxplot-results}
\end{figure}

\begin{figure}
    \centering
    \renewcommand{\arraystretch}{0}
    \begin{tabular}{c@{}c}
        {\footnotesize airfoil} & {\footnotesize bioav} \\ 
        \includegraphics[width=0.48\textwidth]{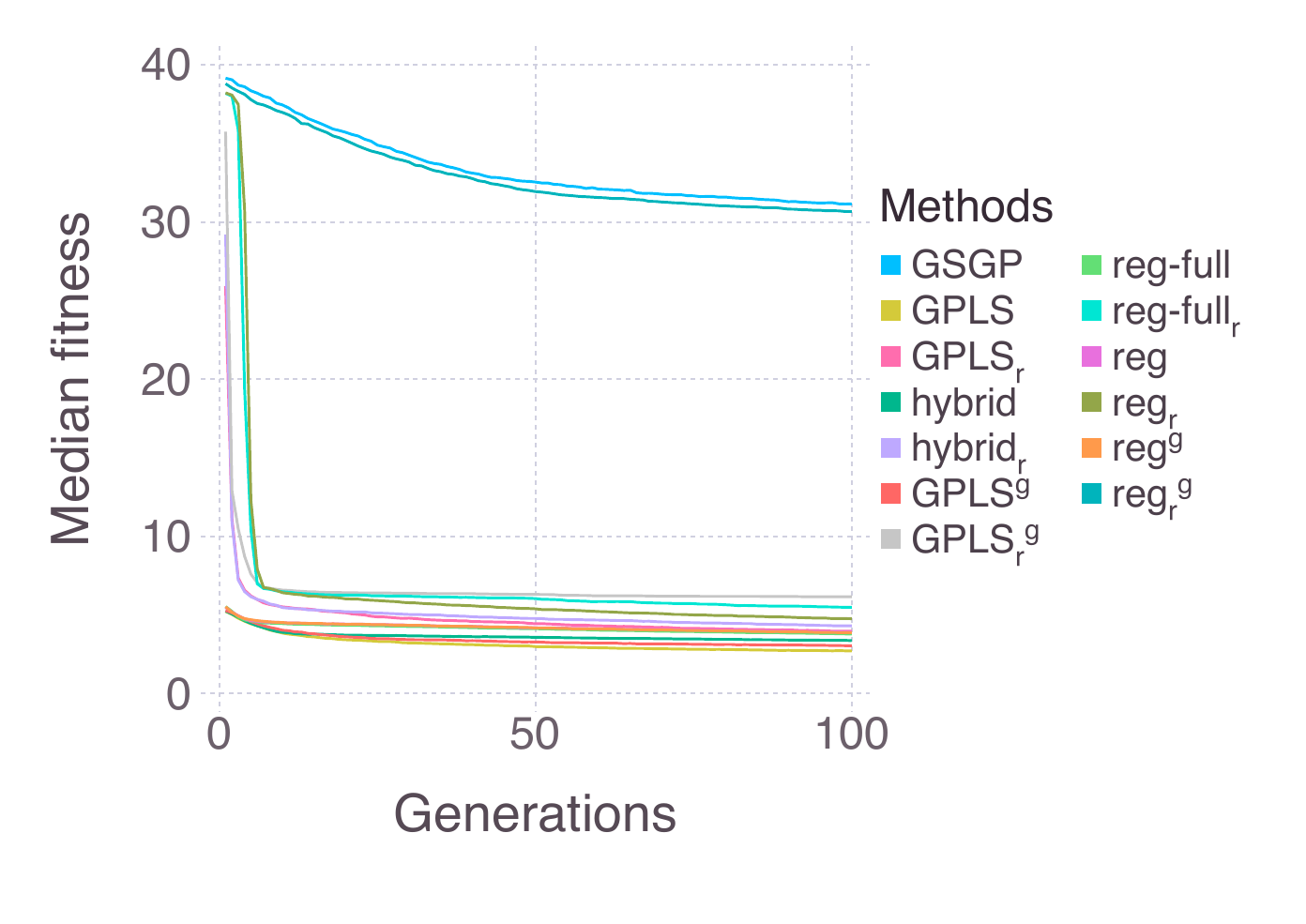} &
        \includegraphics[width=0.48\textwidth]{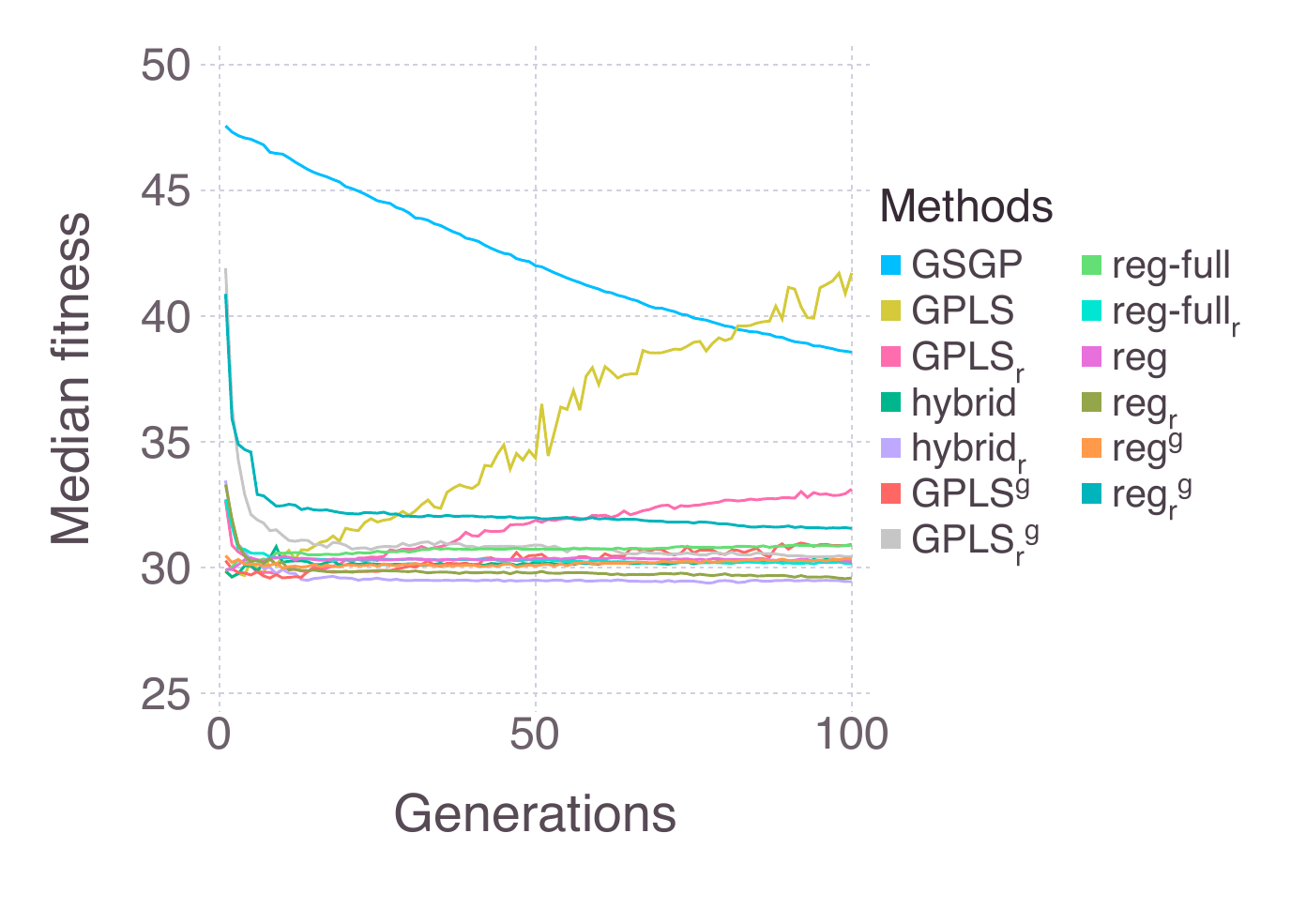} \\
        {\footnotesize concrete} & {\footnotesize parkinson} \\
        \includegraphics[width=0.48\textwidth]{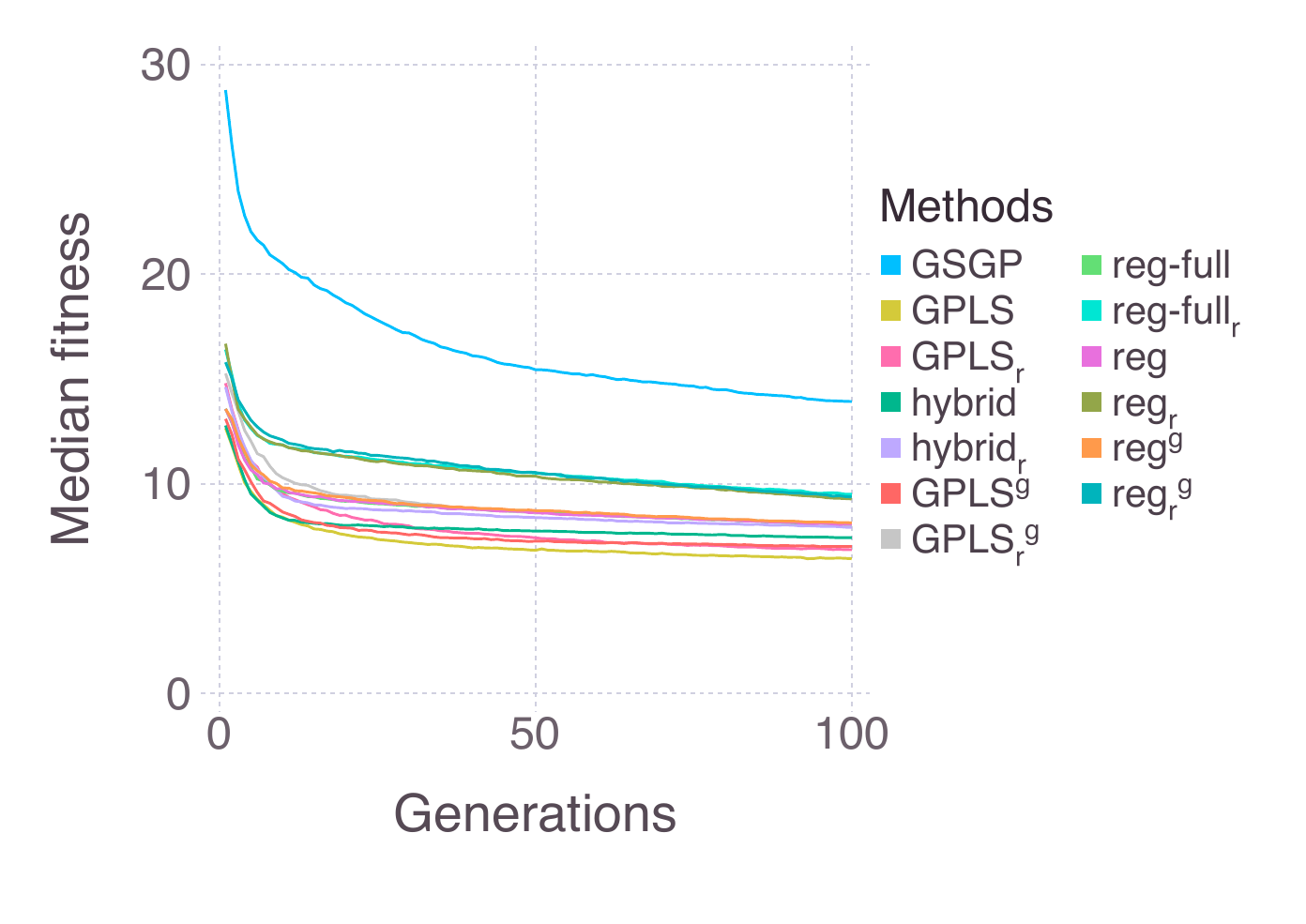} &
        \includegraphics[width=0.48\textwidth]{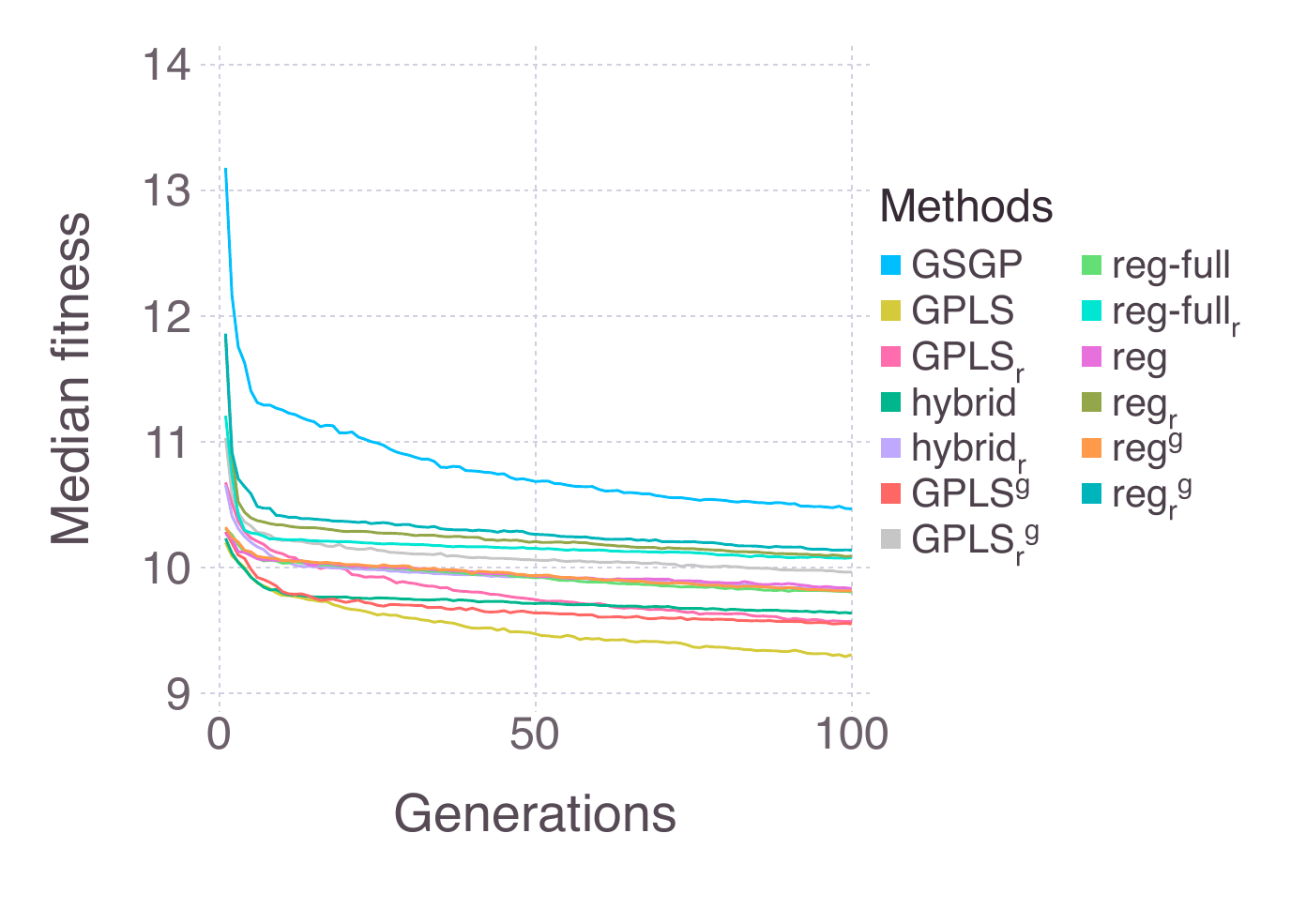} \\
        {\footnotesize ppb} & {\footnotesize slump} \\
        \includegraphics[width=0.48\textwidth]{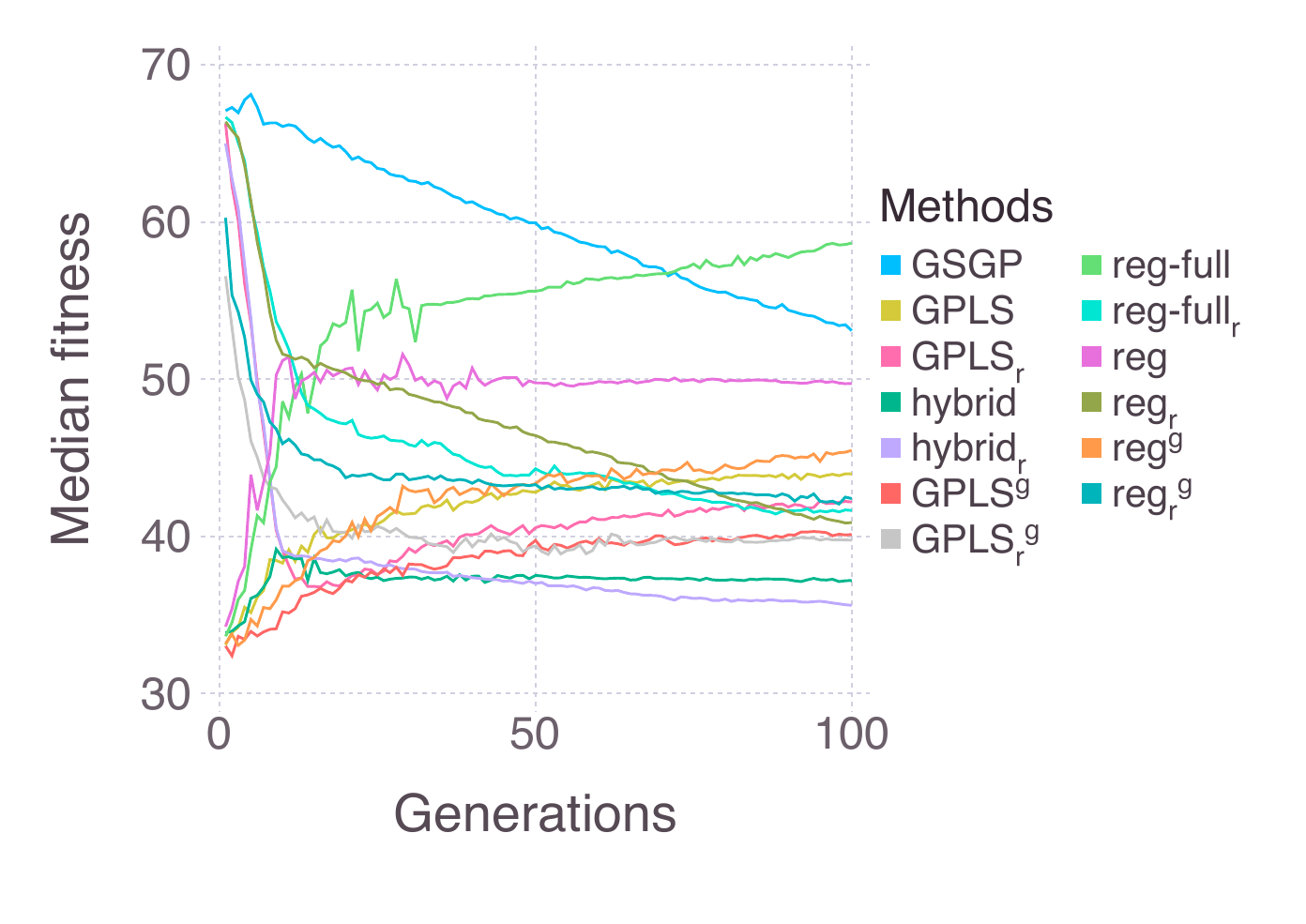} &
        \includegraphics[width=0.48\textwidth]{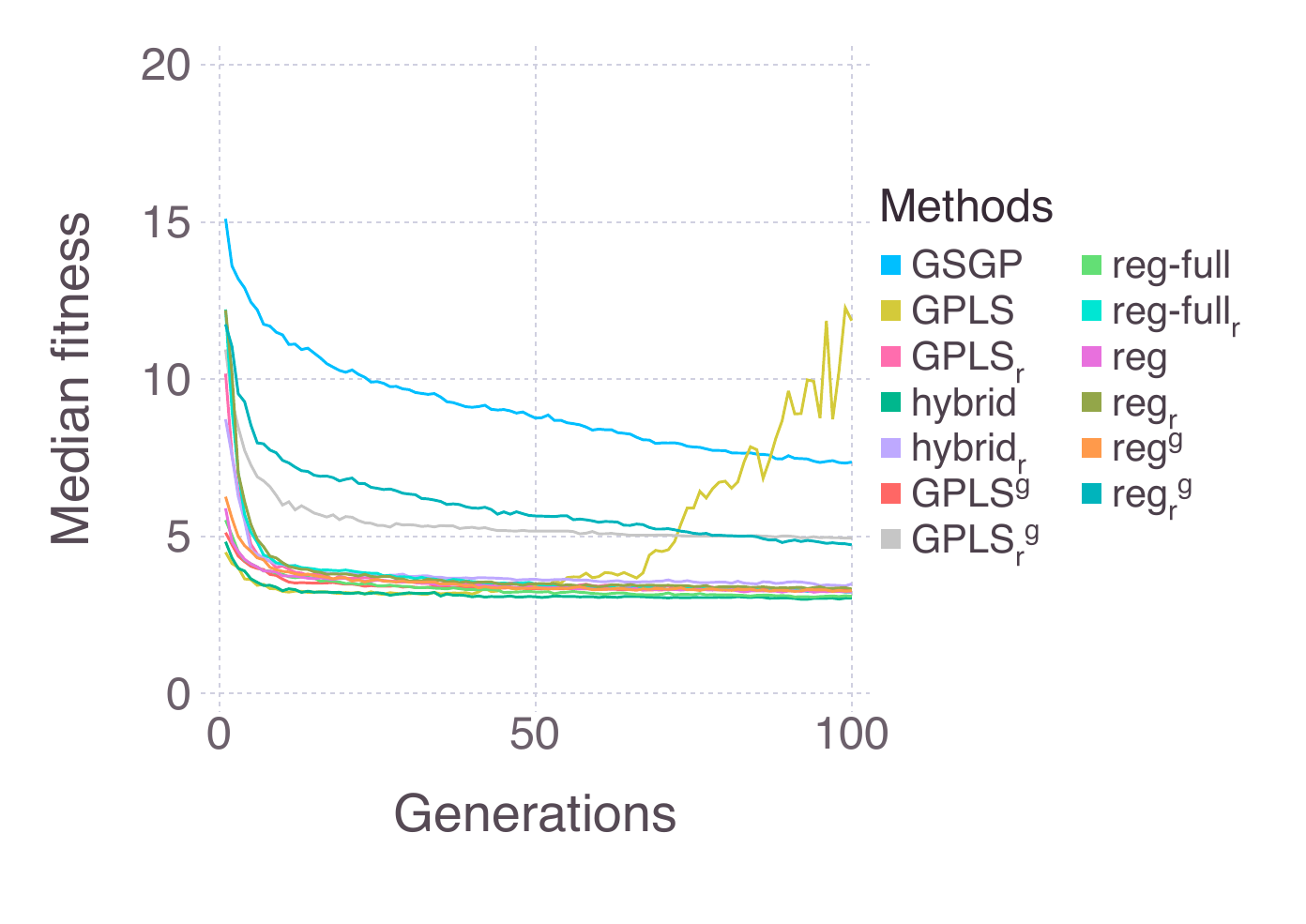} \\
        {\footnotesize LD50} & \\
        \includegraphics[width=0.48\textwidth]{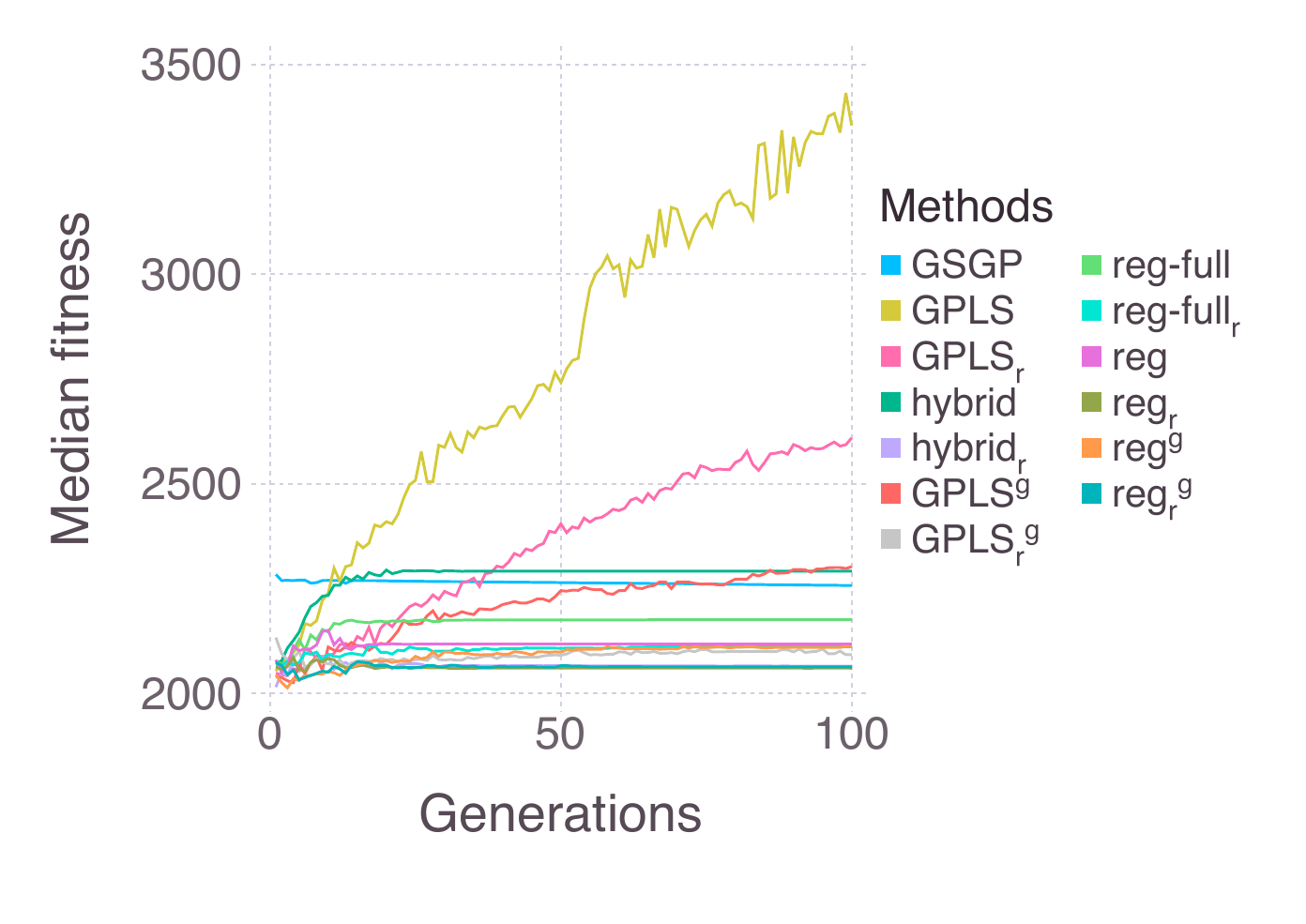} & \\
    \end{tabular}
    \caption{Evolution of the median best fitness across $100$ generations for all considered problems.}
    \label{fig:dynamics-results}
\end{figure}

\begin{figure}
    \centering
    \renewcommand{\arraystretch}{0}
    \begin{tabular}{c@{}c}
        {\footnotesize airfoil} & {\footnotesize bioav} \\ 
        \includegraphics[width=0.47\textwidth]{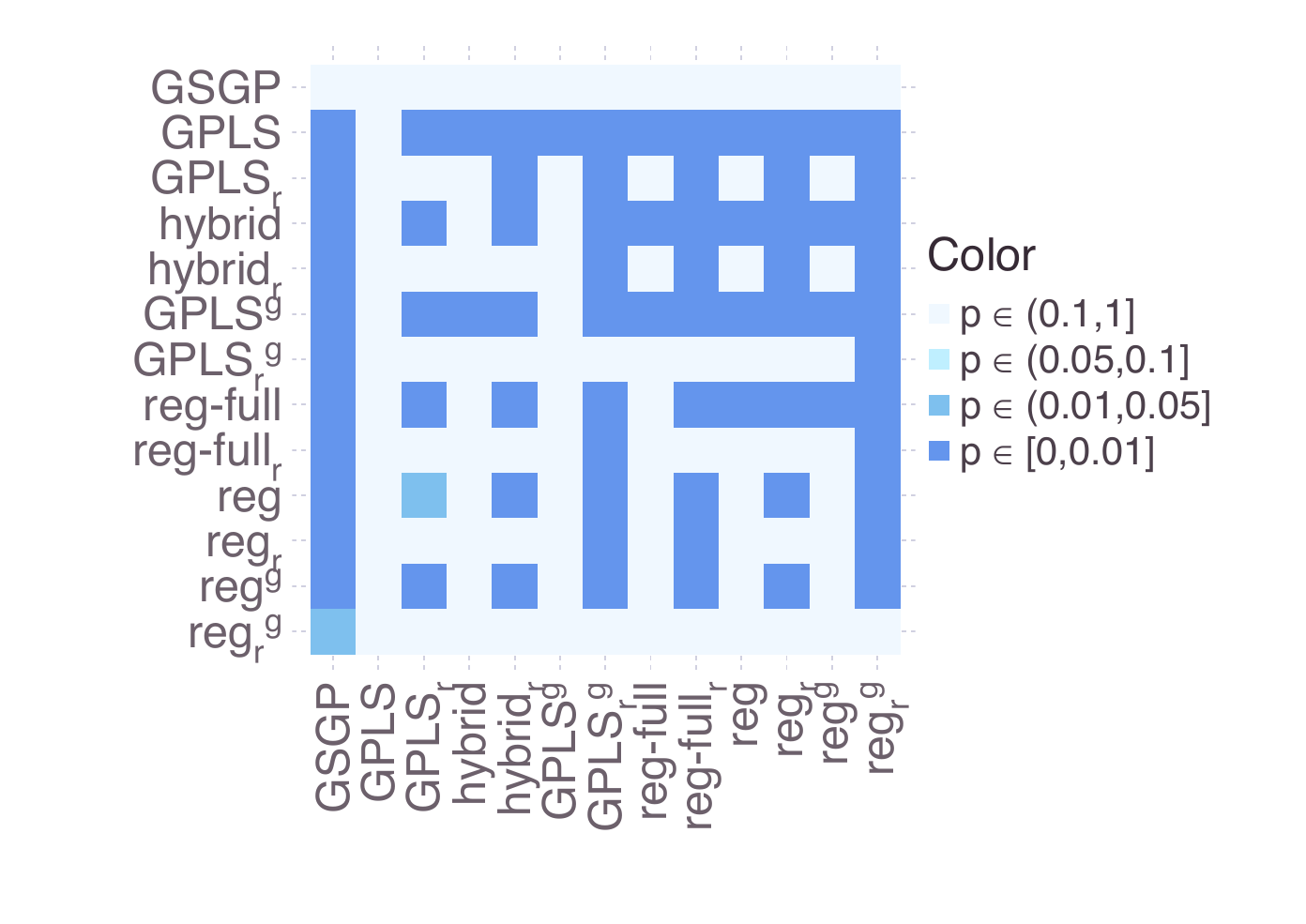} &
        \includegraphics[width=0.47\textwidth]{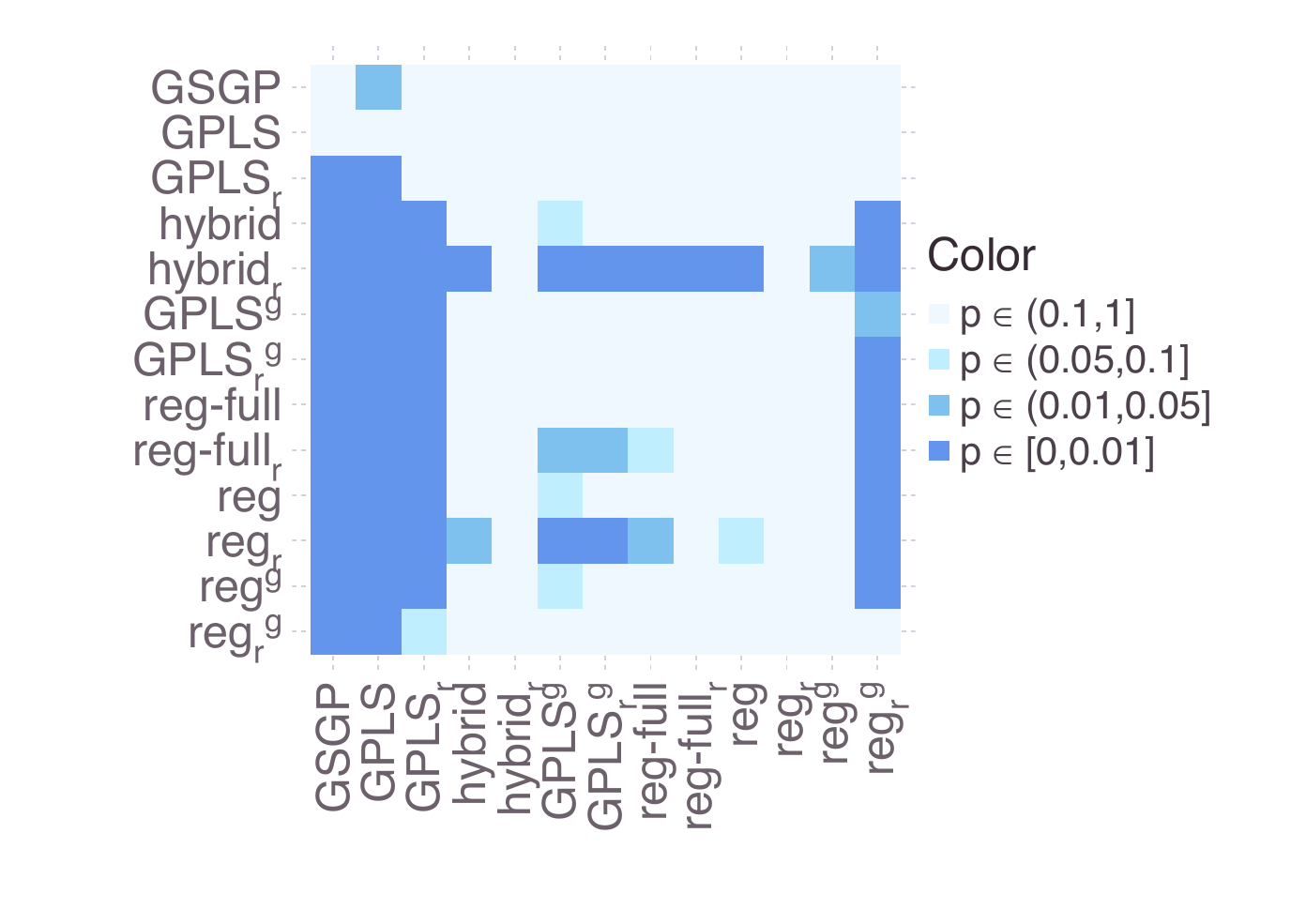} \\
        {\footnotesize concrete} & {\footnotesize parkinson} \\
        \includegraphics[width=0.47\textwidth]{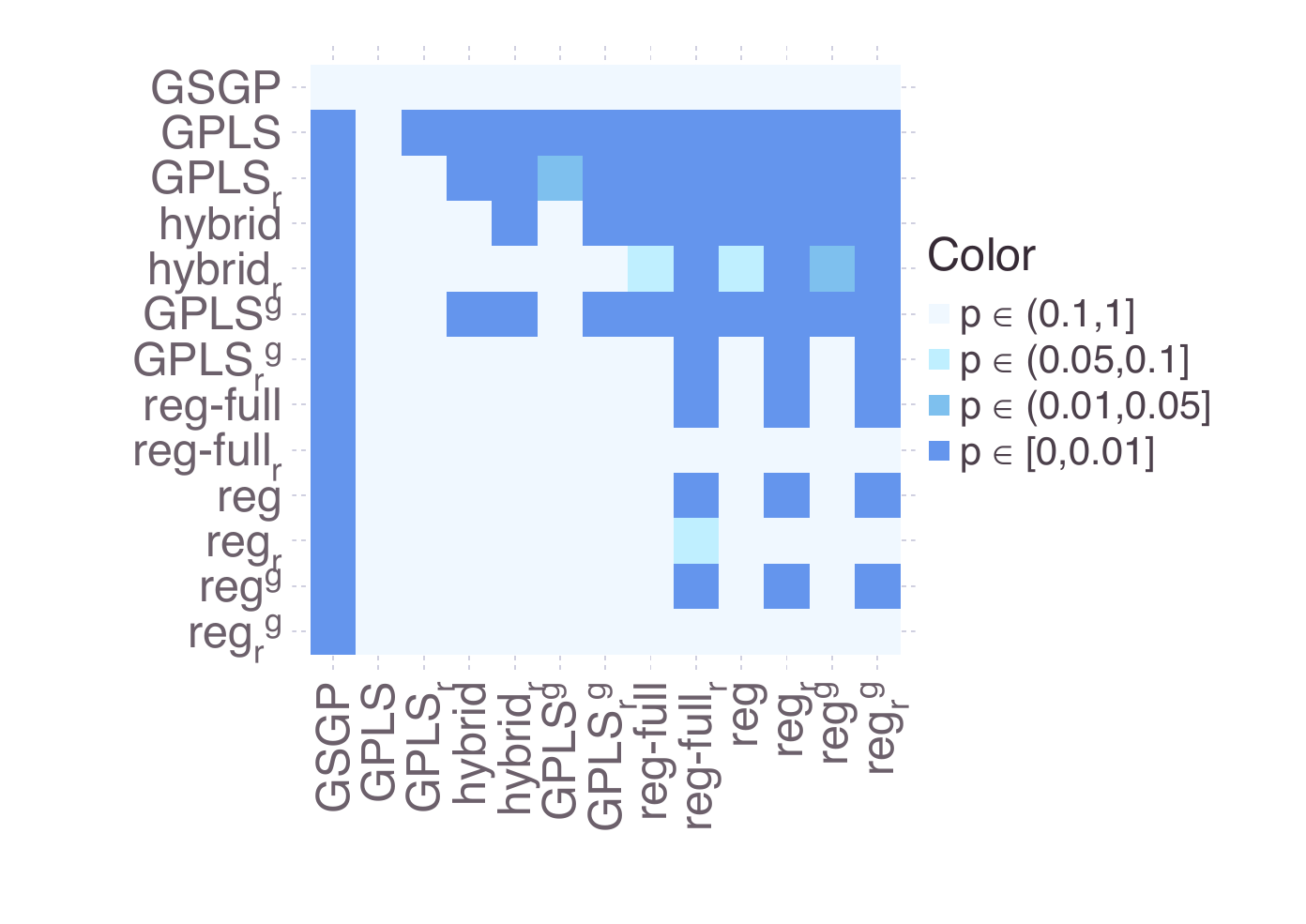} &
        \includegraphics[width=0.47\textwidth]{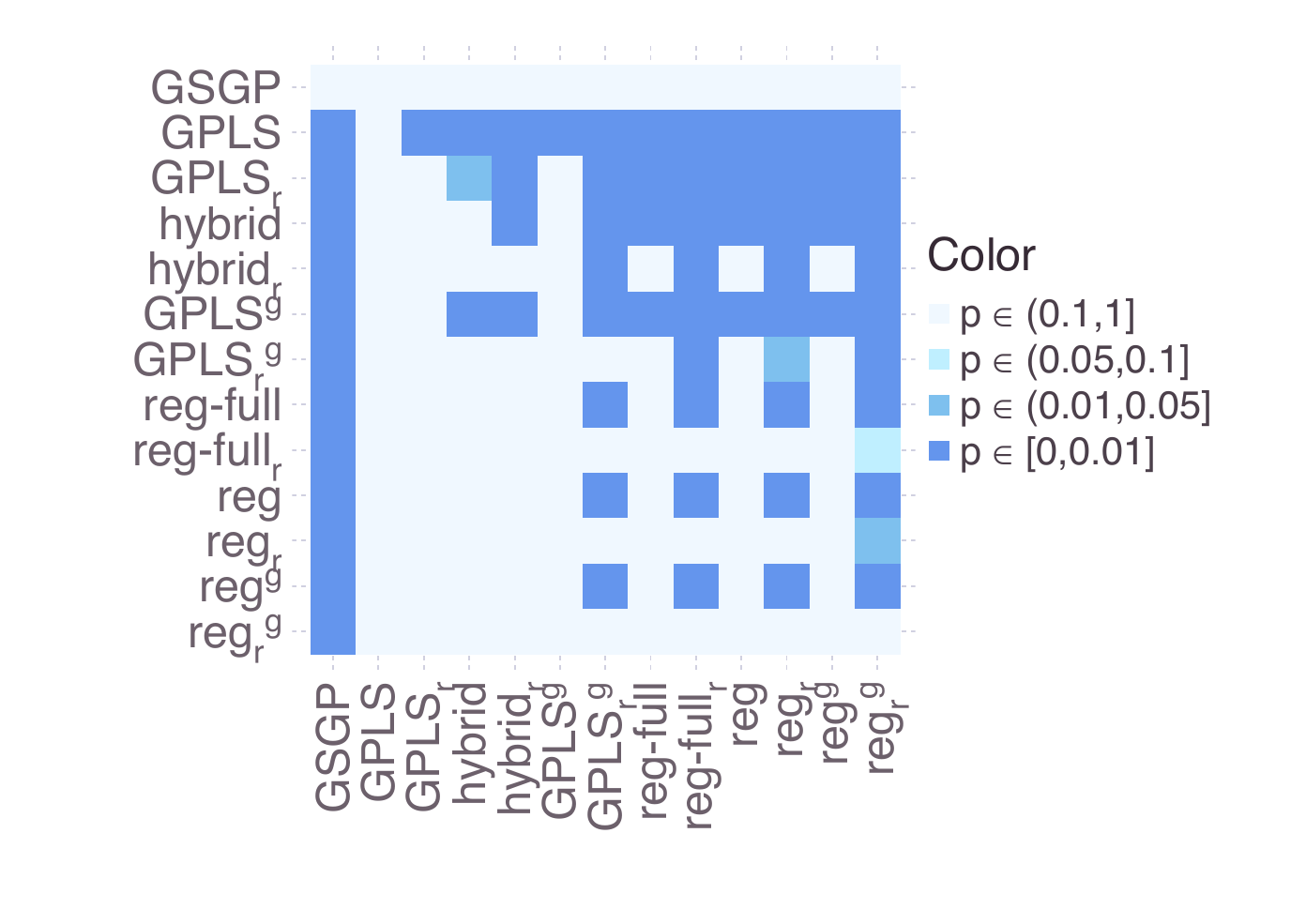} \\
        {\footnotesize ppb} & {\footnotesize slump} \\
        \includegraphics[width=0.47\textwidth]{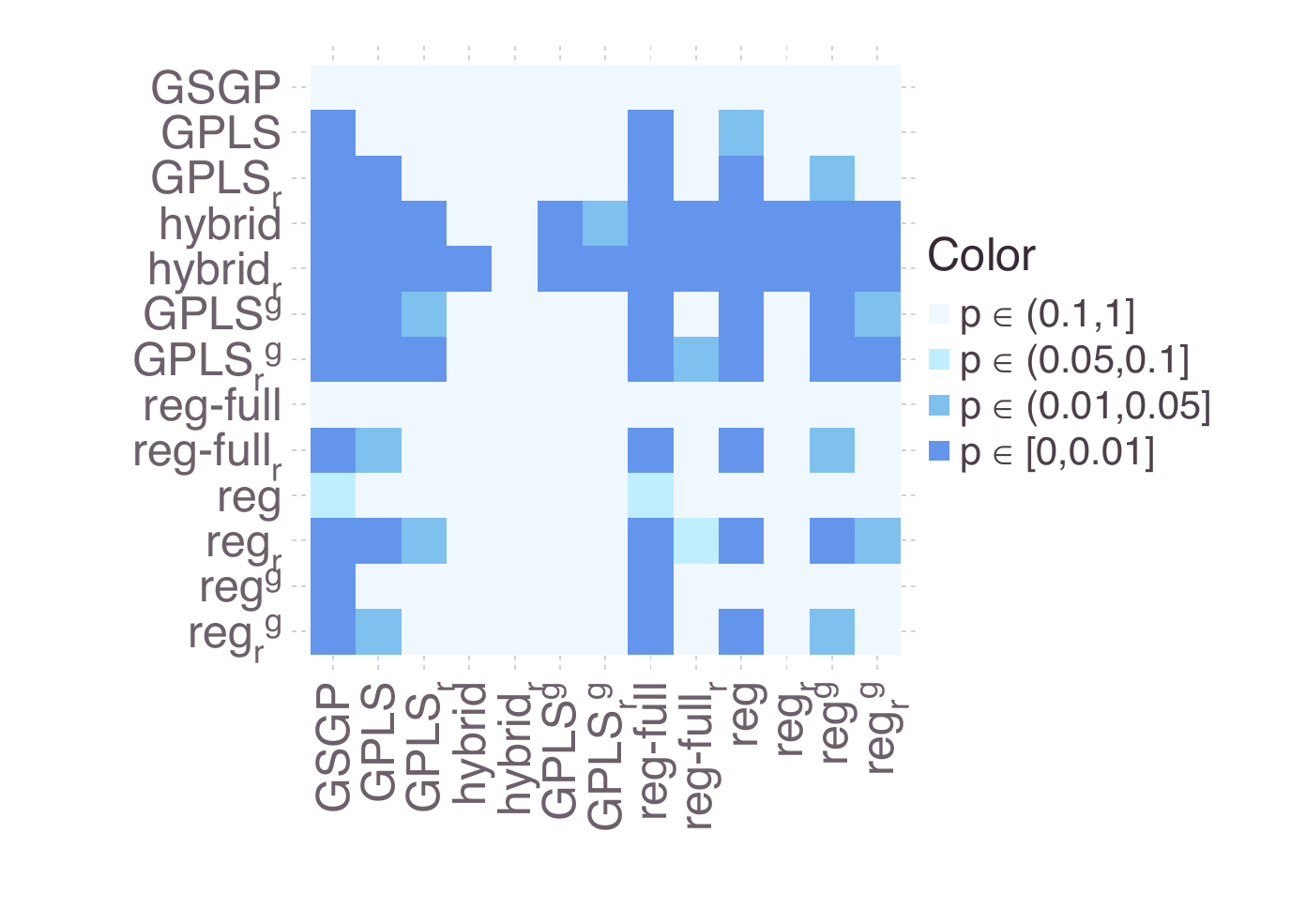} &
        \includegraphics[width=0.47\textwidth]{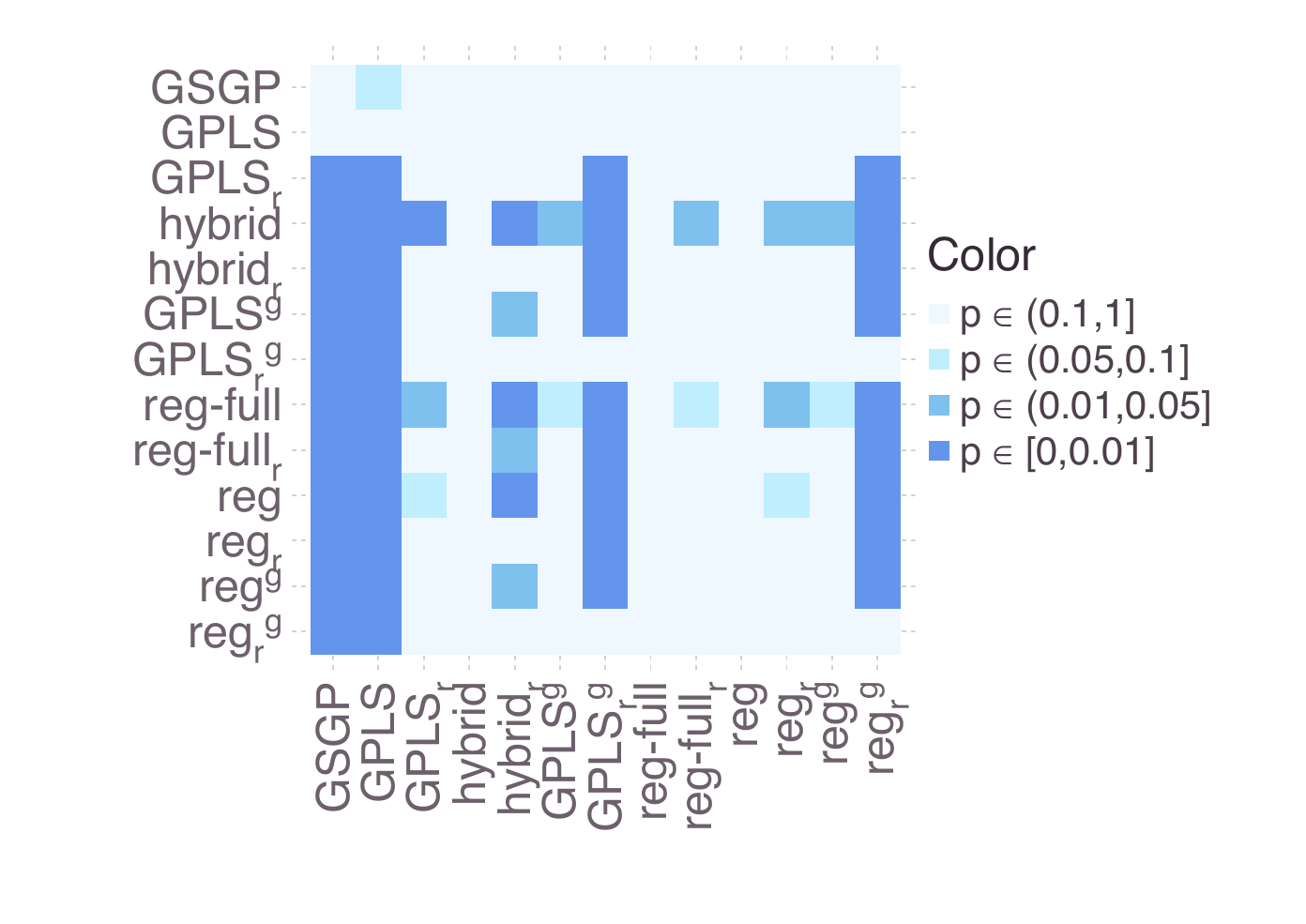} \\
        {\footnotesize LD50} & \\
        \includegraphics[width=0.47\textwidth]{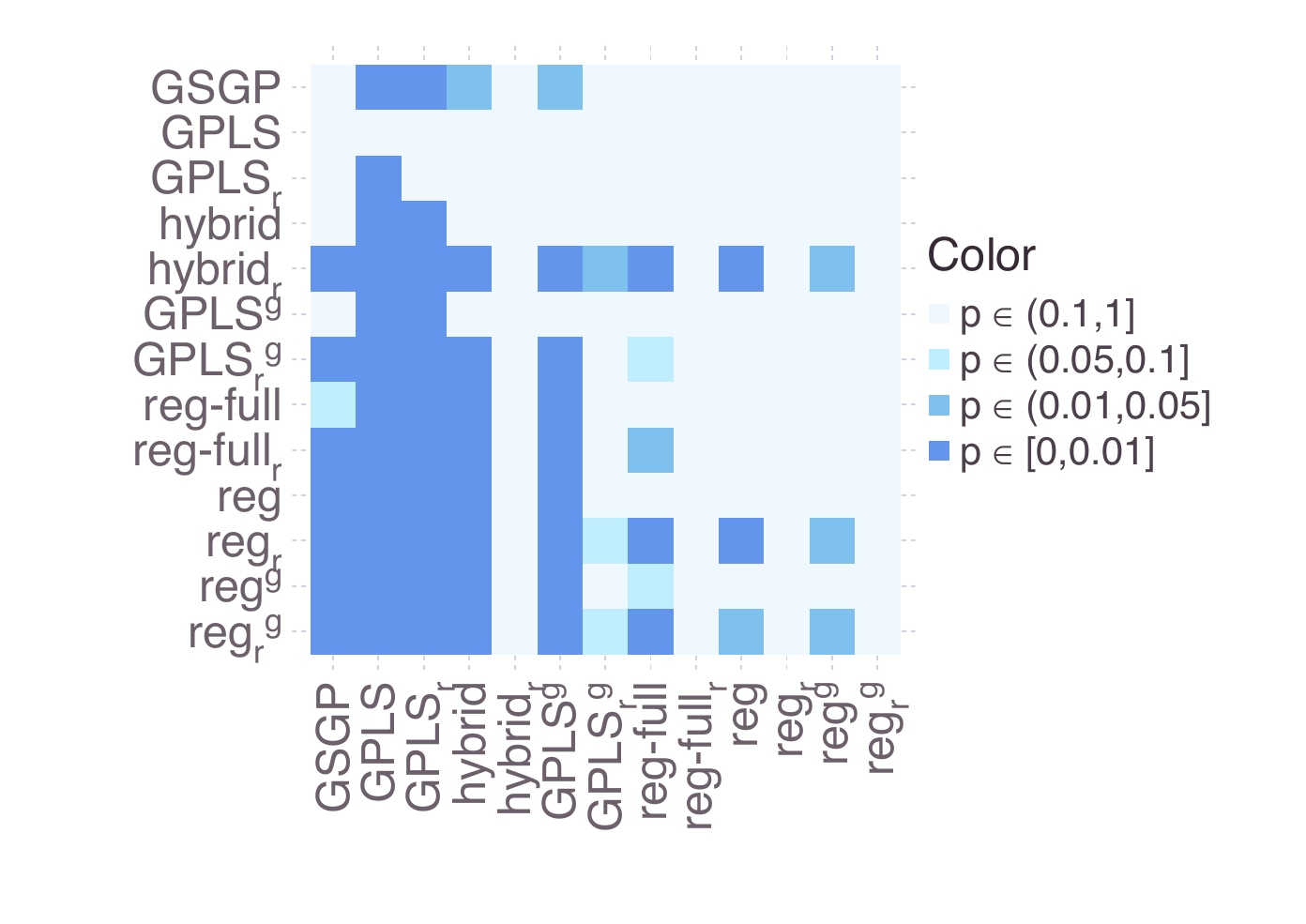} & \\
    \end{tabular}
    \caption{Statistical comparison of the tested algorithms via a one-tailed Mann-Whitney U-test. The alternative hypothesis is that the first compared algorithm (in the rows) produces results with better fitness than the second algorithm (in the columns) more than $1/2$ of the times.}
    \label{fig:stat-results}
\end{figure}

\begin{figure}
    \centering
    \renewcommand{\arraystretch}{0}
    \begin{tabular}{c@{}c}
        {\footnotesize airfoil} & {\footnotesize bioav} \\ 
        \includegraphics[width=0.48\textwidth]{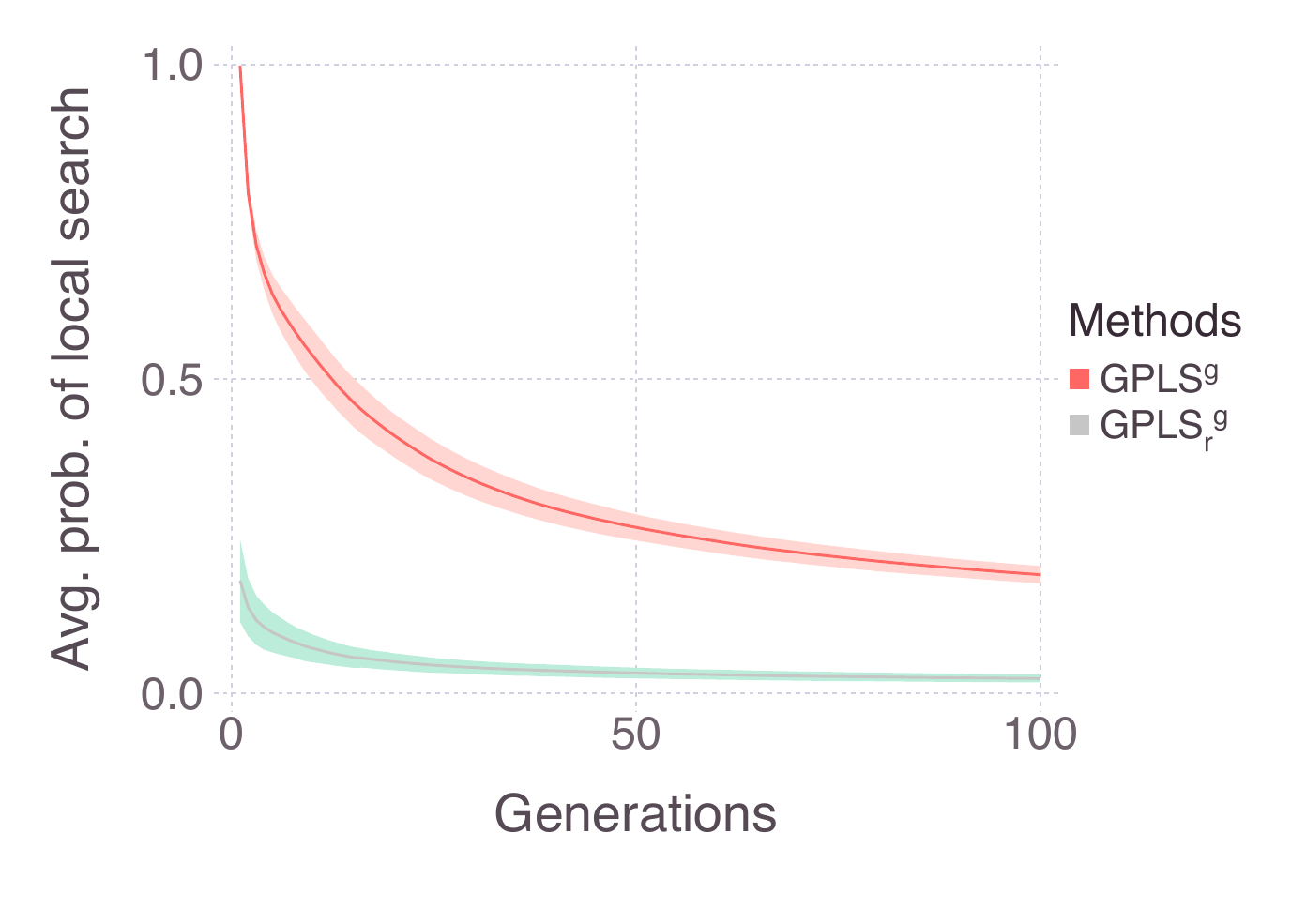} &
        \includegraphics[width=0.48\textwidth]{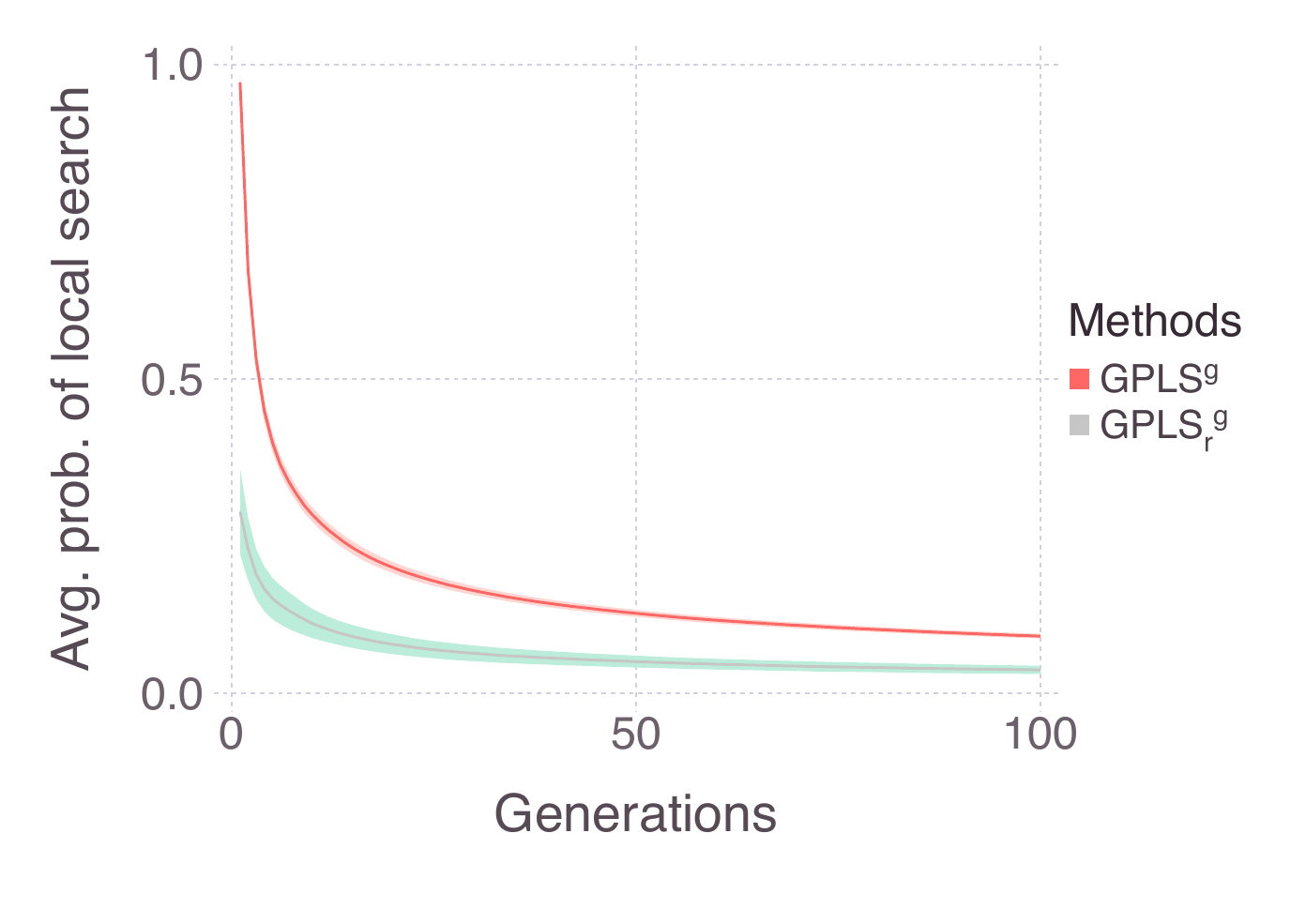} \\
        {\footnotesize concrete} & {\footnotesize parkinson} \\
        \includegraphics[width=0.48\textwidth]{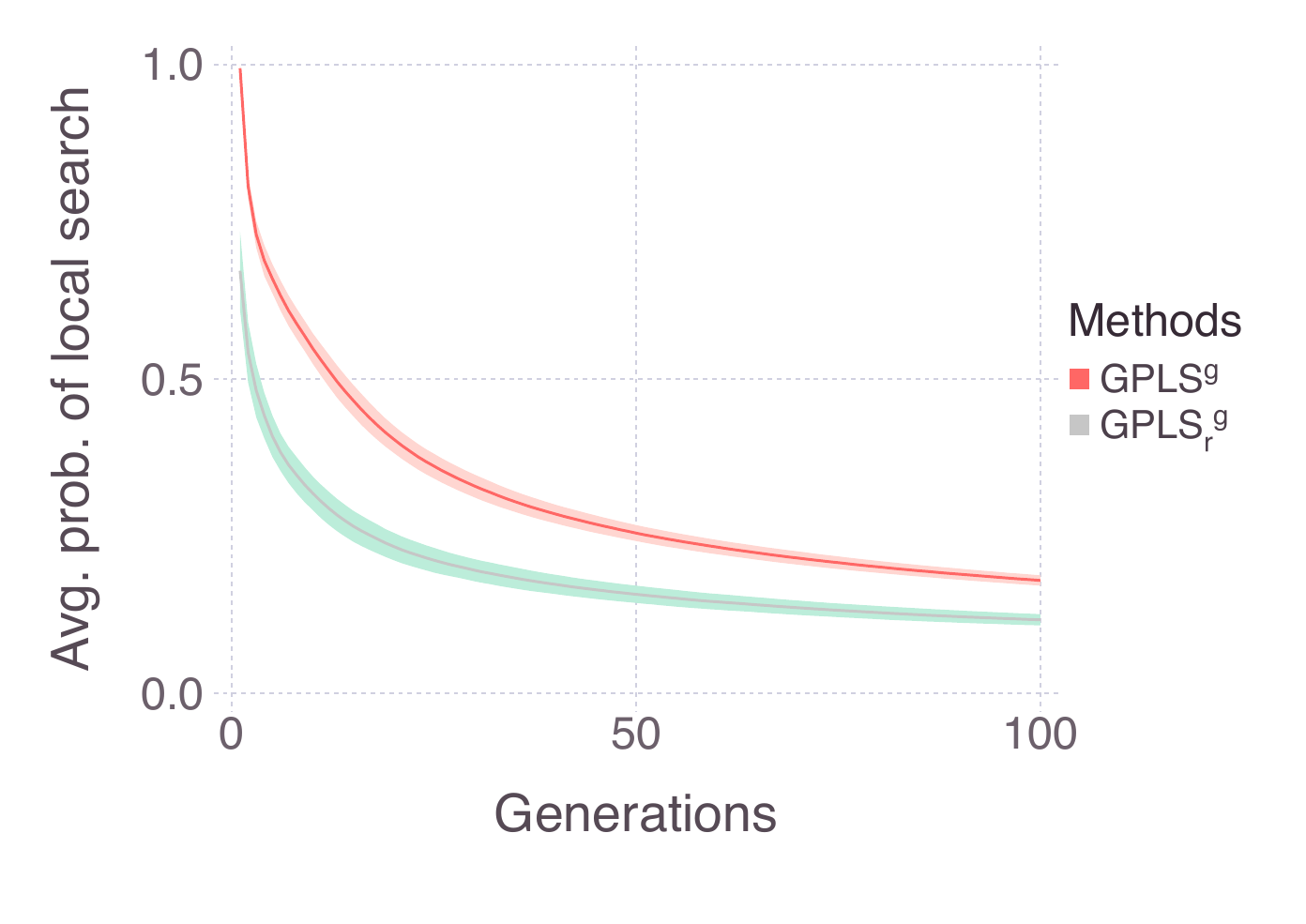} &
        \includegraphics[width=0.48\textwidth]{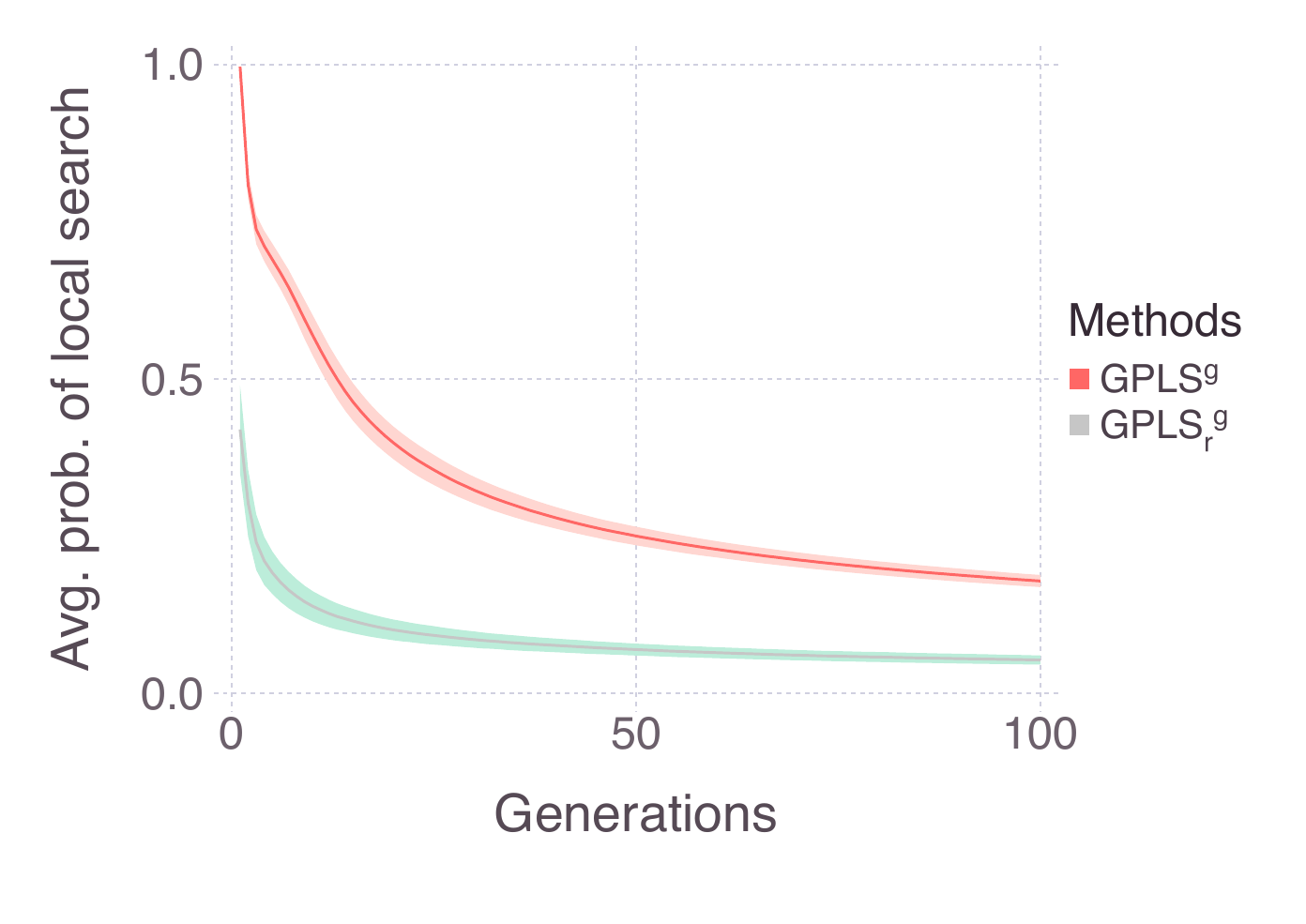} \\
        {\footnotesize ppb} & {\footnotesize slump} \\
        \includegraphics[width=0.48\textwidth]{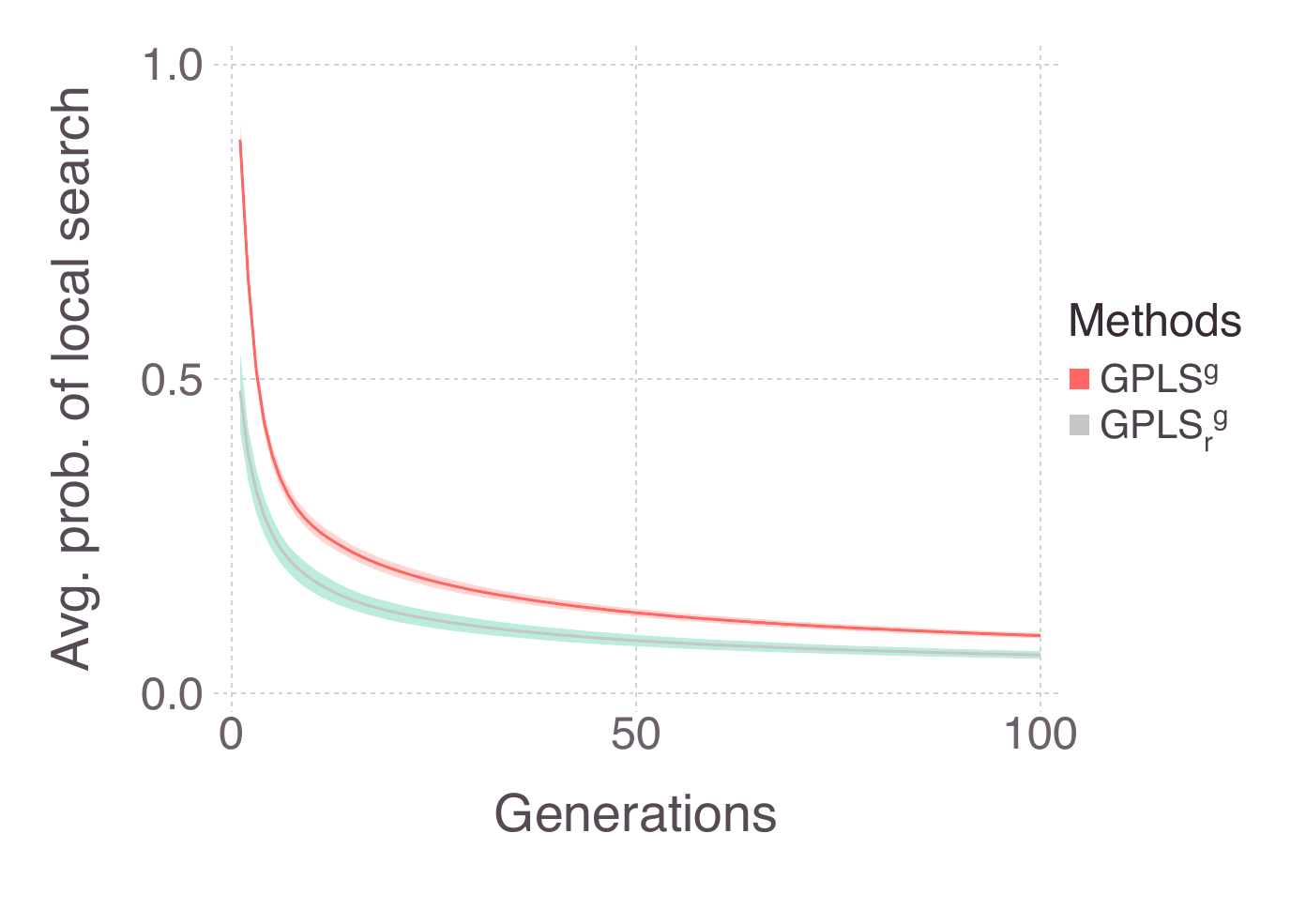} &
        \includegraphics[width=0.48\textwidth]{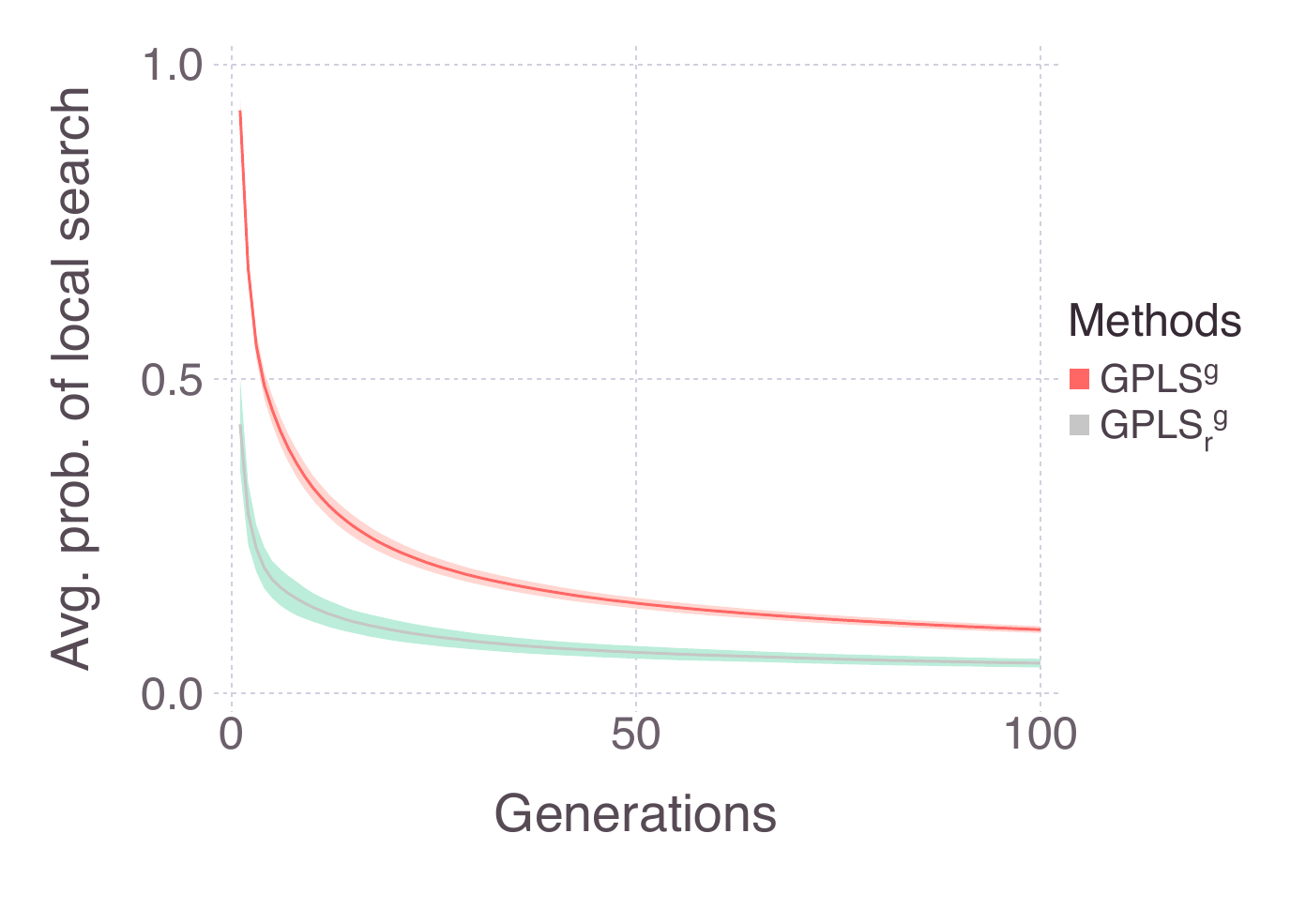} \\
        {\footnotesize LD50} & \\
        \includegraphics[width=0.48\textwidth]{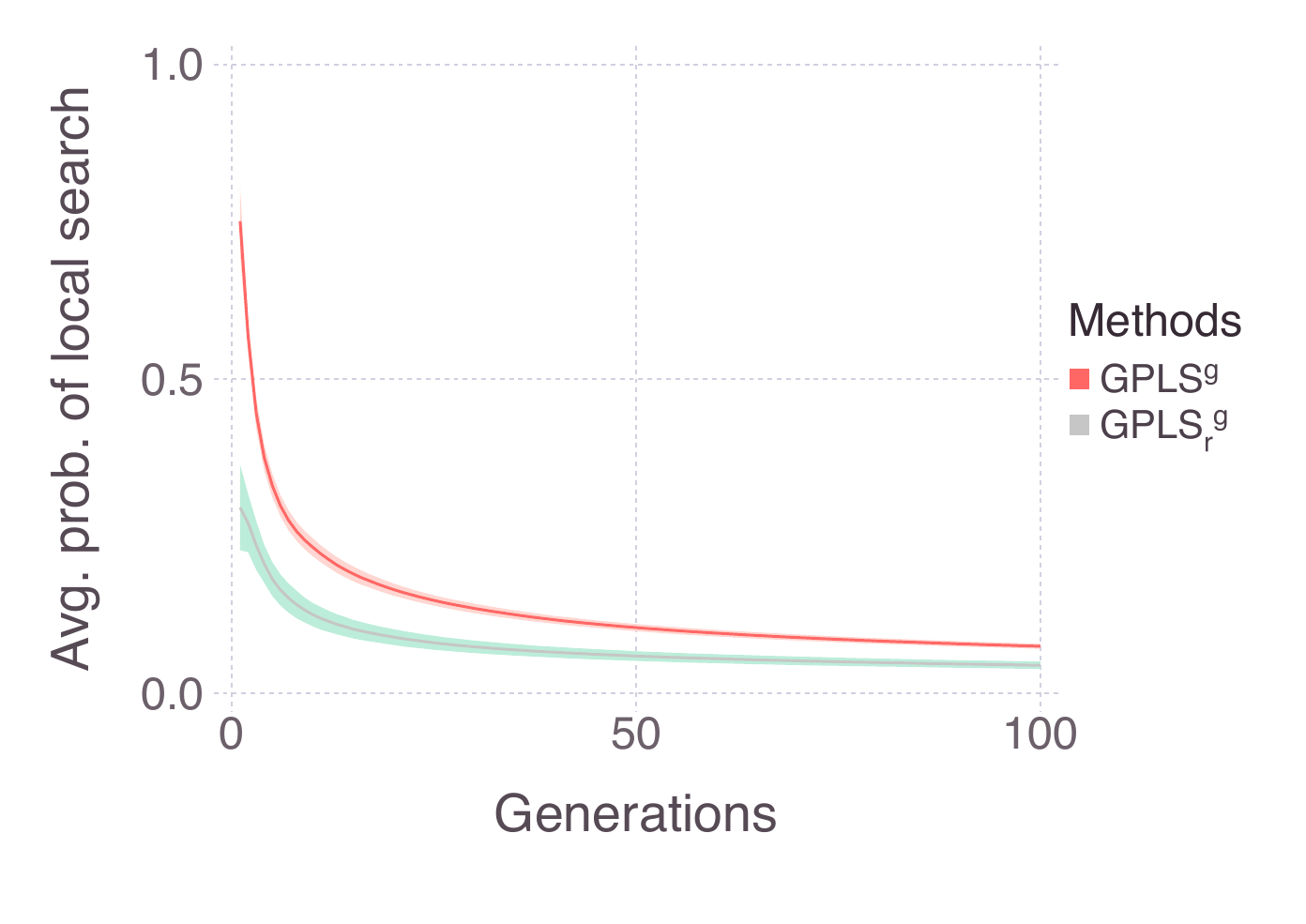} & \\
    \end{tabular}
    \caption{The evolution of the average probability of applying the local search for all methods using the \gen algorithm where local search can be applied for all generations. The shadow represents one standard deviation above and below the average.}
    \label{fig:LS-probability}
\end{figure}

The fitness distribution on the test set at the last generation is depicted in Fig.~\ref{fig:boxplot-results}. The evolution of the median best fitness, also on the test set, computed across all runs with respect to the number of generations, is reported in Fig.~\ref{fig:dynamics-results}. The statistical comparison of the fitness obtained at the last generation on the test set is presented in Fig.~\ref{fig:stat-results}. For comparing an algorithm with all others with a statistical significance of $\alpha = 0.01$, it is sufficient to consider the row corresponding to the algorithm and see which cells have the darkest color ($p$-value at most $0.01$).
Given two algorithms $i$ and $j$, a dark blue in position $(i,j)$ means that algorithm $i$ produces better results than algorithm $j$ in a statistically significant manner. Vice versa, if the dark blue color is in position $(j,i)$.
If neither happens, none of the algorithms generated better results than the other in a statistically significant way.
Finally, Fig.~\ref{fig:LS-probability} displays, for all methods making use of the \gen algorithm, the probability of performing a local search step with respect to the number of generations.

By analyzing the results, we can identify three main behaviours:
\begin{itemize}
    \item There is no overfitting, \GPLS is the best performer, and \GSGP is the worst one. This happens for the airfoil, concrete, and parkinson datasets.
    \item There is overfitting, \GPLS is now the worst performer, and most of the other methods can manage overfitting while still outperforming \GSGP. This is the case for the bioav, slump, and LD50 datasets.
    \item The last behaviour appears only in the ppb datasets and shows \GPLS not overfitting but with other methods being the best performers. In this case, some of the competitors overfit.
\end{itemize}
We will discuss the results on each dataset according to the above behaviours.

As it is possible to observe, for the airfoil problem, all of the algorithms produce better results than \GSGP, except for \regrg. While there are variations among the different methods, it is possible to observe that all of them converge rapidly---in less than $10$ generations---to a fitness value below $10$ without any noticeable improvement after. The boxplots and the statistical tests show that \GPLS is the best performer. The overfitting prevention techniques do not provide any advantage here, and, in a limited part, the RMSE tends to be slightly higher than \GPLS for them. However, it is important to remark that, excluding the case of \regrg, all methods are still able to outperform \GSGP in a statistically significant way.

For the concrete dataset, the behaviour is similar to the one on airfoil, with \GPLS being the best performer, also in a statistically significant way. It is possible to observe that the \reg-based methods perform worse than the ones using GSM-LS, and that ridge regression reduces the ability to find solutions with a better fitness. As with airfoil, there is a quick improvement at the beginning for most algorithms excluding \GSGP, followed in some cases by a slow improvement of the fitness.

A similar but less pronounced pattern appears from the results on the parkinson dataset. \GPLS is still the best performer, with \GSGP being the worst one. The difference here is less pronounced, but the main characteristics found in the results on the concrete dataset are still present: the \reg-based methods generally perform worse than the one based on GSM-LS, and the use of ridge regression limits the ability to find solutions with fitness on par with the methods using least squares regression. The change in fitness with time shows a fast convergence to good-quality solutions for most methods, but, differently from the airfoil and concrete datasets, some of the local search methods are still improving after $100$ generations.

The bioav dataset provides a first example where \GPLS is not the best performer. In this case, it is the worst-performing method and significantly worse than plain \GSGP. The \regrg method is among the worst considering the ones that still perform better than \GSGP, similar to what happened in the airfoil dataset. By observing the evolution of the median fitness over the number of generations, it is interesting to notice not only that \GPLS starts to overfit after only a few generations, but also that \GPLSr is slowly getting worse and worse on the test set. This shows that, at least in this case, the use of ridge regression is not sufficient to counteract the overfitting of \GPLS.

The results on the slump dataset show many similarities with the ones on the bioav dataset, with \GPLS being the worst performer due to overfitting. Similarly to what we observed in other datasets, \regrg converges slowly to good-quality solutions compared to the other local search methods. From the evolution of the fitness with respect to the number of generations, it is possible to observe that the overfitting for \GSGP starts later (after more than $50$ generations) and that no overfitting seems to happen for \GPLSr, as in the case of the bioav dataset.

In the LD50 dataset, it is possible to observe overfitting for most methods based on GSM-LS, with the \reg-based methods improving with respect to \GSGP. An interesting pattern appears when observing the evolution of the median fitness with time. \GPLS and \GPLSr immediately start to overfit, following the same pattern observed in the bioav dataset, with \GPLS getting worse on the test set faster than \GPLSr. Thus, while the use of ridge regression is useful to limit the amount of overfitting, it is not sufficient to completely counteract it, at least in the case of \GPLS and when not combined with other overfitting prevention methods. An interesting remark is that overfitting also appears for \GPLSg, even though at a much slower pace.

The results on the last dataset, ppb, are different from all others, since \GPLS is still performing better than \GSGP. However, by observing the evolution of the fitness over time, it is clear that there is overfitting from the first few generations already. In general, the patterns that can be found are either a slow but constant improvement in fitness (\GSGP and \regr), an initial large improvement in the fitness followed by an almost stationary situation (\regfullr, \regrg, \GPLSrg, and \hybridr), and an almost immediate overfitting (all remaining methods).

If we focus only on the methods using the \gen algorithm where local search can be applied for all generations (i.e., \GPLSg and \GPLSrg) and, in particular, the probability of applying a local search step, it is possible to observe a general trend with a distinction between the use of ridge regression and the ordinary least squares regression.
The general trend is a decline in probability followed by a stabilization or a slower decline. This pattern is consistent with the effect of local search in many datasets: the fitness improves rapidly at the beginning of the evolution, and, subsequently, only small improvements are possible. Hence, one can expect that successful local search steps will be fewer and fewer with time, thus leading to a general decrease in the probability of applying the local search step.
Moreover, the probability decreases faster and stabilizes to smaller values when applying ridge regression. Thus, it seems that the combination of the two strategies on local search makes their contribution marginal in the evolutionary process, i.e., most local search steps are performed at the beginning.

As it is possible to observe from the results, there is no clear method outperforming all the others. Concerning the research questions that we investigated, the main conclusions that can be drawn are the following:
\begin{itemize}
    \item[\textbf{RQ1}] Most overfitting prevention strategies are effective, and overfitting can be avoided in most cases. In particular, ridge regression can reduce the amount of overfitting of \GPLS without eliminating it. On the other hand, the early stopping of the local search is effective in most cases, but can still lead to large overfitting (e.g., in the LD50 dataset). The use of \gen appears to be the more consistent in preventing overfitting, even considering the slight decrease in performance in one of the datasets (ppb for the \regg method).
    \item[\textbf{RQ2}] Most overfitting prevention strategies provide consistently better performances than \GSGP. Thus, even if they can hinder the effect of local search in datasets not prone to overfitting, the final effect is an increase in performance compared to \GSGP and a reduction of the tendency to overfit.
    \item[\textbf{RQ3}] No strategy is consistently better than all the others. When there is no overfitting, plain \GPLS can perform extremely well. All overfitting prevention strategies work (with caveats for some datasets, as discussed above), and can be combined, even if the combination of \gen and ridge regression seems not to give the best situation.
\end{itemize}

In summary, is it worthwhile to apply a local search step in \GSGP? Yes, we can consistently obtain better fitness for some of the algorithms. Can we limit the risk of overfitting easily? Yes, there are multiple overfitting prevention strategies. Is there one preferred method to limit overfitting? While \reg appears to be the most consistent one, it is not a clear winner. It is fundamental to have one, or a combination of more overfitting prevention strategies, but which one to adopt is not that essential compared to the presence of such a strategy.

\section{Conclusions}
\label{sec:conclusions}

In this work, we performed a comprehensive study of local search in \GSGP, comparing the two approaches of \GPLS and \reg and assessing their performance with different overfitting prevention strategies. The strategies include the application of local search only to the first $10$ generations, the use of ridge regression in the local step, and an adaptive algorithm \gen.
The results show that with a local search strategy, it is possible to outperform \GSGP. Additionally, an overfitting prevention strategy is essential to avoid the risk of overfitting inherent in the considered local search methods. Interestingly, there is no clear ``best'' overfitting prevention strategy, even if the ones based on the \gen algorithm perform more consistently.

In future work, it would be interesting to study the relation between overfitting and moves in the semantic space of the individuals, to discover whether the ``path'' an individual follows in the semantic space is indicative of overfitting. This analysis would help the design of better overfitting prevention strategies. Finally, more effective local search algorithms can be defined. However, in this case, it would be necessary to consider the potential additional computational effort. While this was not the case for the studied algorithms---the local search is not computationally expensive---it could be a relevant point for computationally expensive local search steps.

\bibliographystyle{abbrv}
\bibliography{biblio}

\begin{thebibliography}{10}

\bibitem{archetti2007genetic}
F.~Archetti, S.~Lanzeni, E.~Messina, and L.~Vanneschi.
\newblock Genetic programming for computational pharmacokinetics in drug
  discovery and development.
\newblock {\em Genetic Programming and Evolvable Machines}, 8:413--432, 2007.

\bibitem{azad2014simple}
R.~M.~A. Azad and C.~Ryan.
\newblock A simple approach to lifetime learning in genetic programming-based
  symbolic regression.
\newblock {\em Evolutionary computation}, 22(2):287--317, 2014.

\bibitem{brooks1989airfoil}
T.~F. Brooks, D.~S. Pope, and M.~A. Marcolini.
\newblock Airfoil self-noise and prediction.
\newblock Technical report, 1989.

\bibitem{castelli2017evolutionary}
M.~Castelli, I.~Gon{\c{c}}alves, L.~Trujillo, and A.~Popovi{\v{c}}.
\newblock An evolutionary system for ozone concentration forecasting.
\newblock {\em Information Systems Frontiers}, 19:1123--1132, 2017.

\bibitem{CASTELLI2019100313}
M.~Castelli and L.~Manzoni.
\newblock {GSGP-C++} 2.0: A geometric semantic genetic programming framework.
\newblock {\em SoftwareX}, 10:100313, 2019.

\bibitem{EPIA}
M.~Castelli, L.~Manzoni, L.~Mariot, and M.~Saletta.
\newblock Extending local search in geometric semantic genetic programming.
\newblock In {\em Progress in Artificial Intelligence: 19th EPIA Conference on
  Artificial Intelligence, EPIA 2019, Vila Real, Portugal, September 3--6,
  2019, Proceedings, Part I 19}, pages 775--787. Springer, 2019.

\bibitem{castelli2015c++}
M.~Castelli, S.~Silva, and L.~Vanneschi.
\newblock A {C}++ framework for geometric semantic genetic programming.
\newblock {\em Genetic Programming and Evolvable Machines}, 16(1):73--81, 2015.

\bibitem{castelli2017predicting}
M.~Castelli, R.~Sormani, L.~Trujillo, and A.~Popovi{\v{c}}.
\newblock Predicting per capita violent crimes in urban areas: an artificial
  intelligence approach.
\newblock {\em Journal of Ambient Intelligence and Humanized Computing},
  8:29--36, 2017.

\bibitem{castelli2015energy}
M.~Castelli, L.~Trujillo, and L.~Vanneschi.
\newblock Energy consumption forecasting using semantic-based genetic
  programming with local search optimizer.
\newblock {\em Computational intelligence and neuroscience}, 2015:57, 2015.

\bibitem{castelli2015prediction}
M.~Castelli, L.~Trujillo, L.~Vanneschi, and A.~Popovi{\v{c}}.
\newblock Prediction of energy performance of residential buildings: A genetic
  programming approach.
\newblock {\em Energy and Buildings}, 102:67--74, 2015.

\bibitem{castelli2015geometric}
M.~Castelli, L.~Trujillo, L.~Vanneschi, S.~Silva, et~al.
\newblock Geometric semantic genetic programming with local search.
\newblock In {\em Proceedings of the 2015 Annual Conference on Genetic and
  Evolutionary Computation}, pages 999--1006. ACM, 2015.

\bibitem{castelli2016semantic}
M.~Castelli, L.~Vanneschi, L.~Manzoni, and A.~Popovi{\v{c}}.
\newblock Semantic genetic programming for fast and accurate data knowledge
  discovery.
\newblock {\em Swarm and Evolutionary Computation}, 26:1--7, 2016.

\bibitem{castelli2013prediction}
M.~Castelli, L.~Vanneschi, and S.~Silva.
\newblock Prediction of high performance concrete strength using genetic
  programming with geometric semantic genetic operators.
\newblock {\em Expert Systems with Applications}, 40(17):6856--6862, 2013.

\bibitem{chen2011multi}
X.~Chen, Y.-S. Ong, M.-H. Lim, and K.~C. Tan.
\newblock A multi-facet survey on memetic computation.
\newblock {\em IEEE Transactions on Evolutionary Computation}, 15(5):591--607,
  2011.

\bibitem{vcrepinvsek2013exploration}
M.~{\v{C}}repin{\v{s}}ek, S.-H. Liu, and M.~Mernik.
\newblock Exploration and exploitation in evolutionary algorithms: A survey.
\newblock {\em ACM computing surveys (CSUR)}, 45(3):1--33, 2013.

\bibitem{emigdio2014evaluating}
Z.~Emigdio, L.~Trujillo, O.~Sch{\"u}tze, P.~Legrand, et~al.
\newblock Evaluating the effects of local search in genetic programming.
\newblock In {\em EVOLVE-A Bridge between Probability, Set Oriented Numerics,
  and Evolutionary Computation V}, pages 213--228. Springer, Cham, 2014.

\bibitem{enriquez2017automatic}
J.~Enr{\'\i}quez-Z{\'a}rate, L.~Trujillo, S.~de~Lara, M.~Castelli, Z.~Emigdio,
  L.~Mu{\~n}oz, A.~Popovi{\v{c}}, et~al.
\newblock Automatic modeling of a gas turbine using genetic programming: An
  experimental study.
\newblock {\em Applied Soft Computing}, 50:212--222, 2017.

\bibitem{eskridge2004imitating}
B.~E. Eskridge and D.~F. Hougen.
\newblock Imitating success: A memetic crossover operator for genetic
  programming.
\newblock In {\em Proceedings of the 2004 Congress on Evolutionary Computation
  (IEEE Cat. No. 04TH8753)}, volume~1, pages 809--815. IEEE, 2004.

\bibitem{hajek2019forecasting}
P.~Hajek, R.~Henriques, M.~Castelli, and L.~Vanneschi.
\newblock Forecasting performance of regional innovation systems using
  semantic-based genetic programming with local search optimizer.
\newblock {\em Computers \& Operations Research}, 106:179--190, 2019.

\bibitem{kingma2017adam}
D.~P. Kingma and J.~Ba.
\newblock Adam: A method for stochastic optimization.
\newblock {\em arXiv preprint arXiv:1412.6980}, 2014.

\bibitem{koza1994genetic}
J.~R. Koza.
\newblock Genetic programming as a means for programming computers by natural
  selection.
\newblock {\em Statistics and computing}, 4:87--112, 1994.

\bibitem{koza2010human}
J.~R. Koza.
\newblock Human-competitive results produced by genetic programming.
\newblock {\em Genetic programming and evolvable machines}, 11:251--284, 2010.

\bibitem{krawiec2009approximating}
K.~Krawiec and P.~Lichocki.
\newblock Approximating geometric crossover in semantic space.
\newblock In {\em Proceedings of the 11th Annual conference on Genetic and
  evolutionary computation}, pages 987--994. ACM, 2009.

\bibitem{krawiec2013locally}
K.~Krawiec and T.~Pawlak.
\newblock Locally geometric semantic crossover: a study on the roles of
  semantics and homology in recombination operators.
\newblock {\em Genetic Programming and Evolvable Machines}, 14:31--63, 2013.

\bibitem{marquardt1975ridge}
D.~W. Marquardt and R.~D. Snee.
\newblock Ridge regression in practice.
\newblock {\em The American Statistician}, 29(1):3--20, 1975.

\bibitem{mcdermott2012genetic}
J.~McDermott, D.~R. White, S.~Luke, L.~Manzoni, M.~Castelli, L.~Vanneschi,
  W.~Jaskowski, K.~Krawiec, R.~Harper, K.~De~Jong, et~al.
\newblock Genetic programming needs better benchmarks.
\newblock In {\em Proceedings of the 14th annual conference on Genetic and
  evolutionary computation}, pages 791--798, 2012.

\bibitem{moraglio2012geometric}
A.~Moraglio, K.~Krawiec, and C.~G. Johnson.
\newblock Geometric semantic genetic programming.
\newblock In {\em International Conference on Parallel Problem Solving from
  Nature}, pages 21--31. Springer, 2012.

\bibitem{moraglio2004topological}
A.~Moraglio and R.~Poli.
\newblock Topological interpretation of crossover.
\newblock In {\em Genetic and Evolutionary Computation Conference}, pages
  1377--1388. Springer, 2004.

\bibitem{neri2012memetic}
F.~Neri and C.~Cotta.
\newblock Memetic algorithms and memetic computing optimization: A literature
  review.
\newblock {\em Swarm and Evolutionary Computation}, 2:1--14, 2012.

\bibitem{nguyen2020memetic}
P.~T.~H. Nguyen and D.~Sudholt.
\newblock Memetic algorithms outperform evolutionary algorithms in multimodal
  optimisation.
\newblock {\em Artificial Intelligence}, 287:103345, 2020.

\bibitem{pietropolli2022combining}
G.~Pietropolli, L.~Manzoni, A.~Paoletti, and M.~Castelli.
\newblock Combining geometric semantic gp with gradient-descent optimization.
\newblock In {\em Genetic Programming: 25th European Conference, EuroGP 2022,
  Held as Part of EvoStar 2022, Madrid, Spain, April 20--22, 2022,
  Proceedings}, pages 19--33. Springer, 2022.

\bibitem{topchy2001faster}
A.~Topchy and W.~F. Punch.
\newblock Faster genetic programming based on local gradient search of numeric
  leaf values.
\newblock In {\em Proceedings of the 3rd Annual Conference on Genetic and
  Evolutionary Computation}, pages 155--162. Morgan Kaufmann Publishers Inc.,
  2001.

\bibitem{trujillo2018local}
L.~Trujillo, Z.~Emigdio, P.~S. Ju{\'a}rez-Smith, P.~Legrand, S.~Silva,
  M.~Castelli, L.~Vanneschi, O.~Sch{\"u}tze, L.~Mu{\~n}oz, et~al.
\newblock Local search is underused in genetic programming.
\newblock In {\em Genetic Programming Theory and Practice XIV}, pages 119--137.
  Springer, Cham, 2018.

\bibitem{tsanas2009accurate}
A.~Tsanas, M.~Little, P.~McSharry, and L.~Ramig.
\newblock Accurate telemonitoring of parkinson’s disease progression by
  non-invasive speech tests.
\newblock {\em Nature Precedings}, pages 1--1, 2009.

\bibitem{vanneschi2014survey}
L.~Vanneschi, M.~Castelli, and S.~Silva.
\newblock A survey of semantic methods in genetic programming.
\newblock {\em Genetic Programming and Evolvable Machines}, 15(2):195--214,
  2014.

\bibitem{white2013better}
D.~R. White, J.~McDermott, M.~Castelli, L.~Manzoni, B.~W. Goldman,
  G.~Kronberger, W.~Ja{\'s}kowski, U.-M. O’Reilly, and S.~Luke.
\newblock Better gp benchmarks: community survey results and proposals.
\newblock {\em Genetic Programming and Evolvable Machines}, 14:3--29, 2013.

\bibitem{yeh1998modeling}
I.-C. Yeh.
\newblock Modeling of strength of high-performance concrete using artificial
  neural networks.
\newblock {\em Cement and Concrete research}, 28(12):1797--1808, 1998.

\bibitem{zhang2004genetic}
M.~Zhang and W.~Smart.
\newblock Genetic programming with gradient descent search for multiclass
  object classification.
\newblock In {\em European Conference on Genetic Programming}, pages 399--408.
  Springer, 2004.

\end{thebibliography}

\end{document}